%% file: ms.tex
\RequirePackage{snapshot}
\documentclass{article}

\usepackage[nonatbib, preprint]{neurips_2021}
\usepackage[numbers]{natbib}

\usepackage[utf8]{inputenc} %
\usepackage[T1]{fontenc}    %
\usepackage{hyperref}       %
\usepackage{url}            %
\usepackage{booktabs}       %
\usepackage{amsfonts}       %
\usepackage{nicefrac}       %
\usepackage{microtype}      %
\usepackage{xcolor}         %
\usepackage{amsmath}        %
\usepackage{mathtools}      %
\usepackage{bbm}            %
\usepackage{multirow}
\usepackage[font=small]{caption}

\usepackage{graphicx}
\usepackage{graphbox}
\usepackage{float}

\usepackage{epsfig}
\usepackage{placeins} %
\usepackage{multirow}
\usepackage[export]{adjustbox}

\usepackage{graphbox} %
\usepackage{makecell}
\usepackage{pifont}%

\newcommand{\cmark}{\textcolor{green}{\ding{51}}}%
\newcommand{\xmark}{\textcolor{red}{\ding{55}}}%

\usepackage{breqn}

\title{ImageBART: Bidirectional Context with Multinomial Diffusion for Autoregressive Image Synthesis}

\author{%
  Patrick Esser\thanks{The first three authors contributed equally to this
  work.} \qquad Robin Rombach$^*$ \qquad Andreas Blattmann$^*$ \qquad Björn Ommer \\
  Heidelberg Collaboratory for Image Processing, IWR, Heidelberg University, Germany\\
  \url{https://compvis.github.io/imagebart/}
}

\input{figures}

\input{tables}

\input{figures_supplementary}

\input{tables_supplementary}

\DeclareMathOperator*{\argmax}{arg\,max}
\DeclareMathOperator*{\argmin}{arg\,min}

\begin{document}

\maketitle

\begin{abstract}
Autoregressive models and their sequential factorization of the data likelihood
  have recently demonstrated great potential for image representation and
  synthesis. Nevertheless, they incorporate image context in a linear 1D order
  by attending only to previously synthesized image patches above or to the
  left. Not only is this unidirectional, sequential bias of attention unnatural
  for images as it disregards large parts of a scene until synthesis is almost
  complete. It also processes the entire image on a single scale, thus ignoring
  more global contextual information up to the gist of the entire scene.
As a remedy we incorporate a coarse-to-fine hierarchy of context by combining
  the autoregressive formulation with a multinomial diffusion process: Whereas
  a multistage diffusion process successively removes
  information to coarsen an image, we train a (short) Markov chain to invert this
  process. In each stage, the resulting autoregressive ImageBART model
  progressively incorporates context from previous stages in a coarse-to-fine
  manner. 
Experiments show greatly improved image modification capabilities over
  autoregressive models while also providing high-fidelity image generation,
  both of which are enabled through efficient training in a compressed latent space.
  Specifically, our approach can take unrestricted, user-provided masks into
  account to perform local image editing. Thus, 
  in contrast to pure autoregressive models,
  it can solve free-form
  image inpainting and, in the case of conditional models, local, text-guided image
  modification without requiring mask-specific training.
\end{abstract}

\section{Introduction}
Spurred by the increasingly popular attention mechanism, a remarkably simple principle has driven progress in deep generative modeling
over the past few years: Factorizing the likelihood of the data in an
autoregressive (AR) fashion 
\begin{equation}
p(x) = \prod_i p_\theta(x_i \vert x_{<i}) 
\label{eq:ar}
\end{equation}
and subsequently learning the conditional transition probabilities with an expressive neural network such as a transformer \citep{DBLP:conf/nips/VaswaniSPUJGKP17}.  
The success of this approach is evident in domains as diverse as language modeling \citep{DBLP:conf/nips/BrownMRSKDNSSAA20}, music generation \citep{DBLP:journals/corr/abs-2005-00341}, neural machine translation \citep{DBLP:conf/aaai/Li0LZL19, wang2020fairseq}, and (conditional) image synthesis \citep{DBLP:conf/icml/ParmarVUKSKT18, DBLP:conf/icml/ChenRC0JLS20}. 
\enlargethispage{\baselineskip}
However, especially for the latter task of image synthesis, which is also the
focus of this work, the high dimensionality and redundancy present in the data challenges the direct applicability of this approach.

\textbf{Missing Bidirectional Context}
Autoregressive models which represent images as a sequence from the top-left to the bottom-right have demonstrated
impressive performance in sampling novel images and completing the lower half
of a given image \citep{DBLP:conf/icml/ChenRC0JLS20, DBLP:journals/corr/abs-2012-09841}. 
However, the unidirectional, fixed ordering of sequence elements not only imposes a perceptually unnatural bias to attention in images by only considering context information from left or above. It also limits
practical applicability to image modification: Imagine that you only have the lower half of an
image and are looking for a completion of the upper half then these models
fail at this minor variation of the completion task.
The importance of contextual information from both directions
\citep{DBLP:journals/corr/abs-2101-01169} has also been recognized in the context of
language modeling \citep{DBLP:journals/corr/abs-1810-04805, DBLP:conf/acl/LewisLGGMLSZ20}. 
However, simply allowing bidirectional context as in \citep{DBLP:journals/corr/abs-1810-04805} does
not provide a valid factorization of the density function for a generative
model.
Furthermore, the sequential
sampling strategy introduces a gap between training and inference, as
training relies on so-called teacher-forcing
\citep{DBLP:journals/corr/BengioVJS15} (where ground truth is provided
for each step) and inference is performed on previously sampled tokens.
This \emph{exposure bias} can introduce significant accumulations
of errors during the generation process, affecting sample quality and coherence
\citep{DBLP:journals/corr/RanzatoCAZ15}.

\textbf{Global Context \& Control via Multinomial Diffusion}
We propose a coarse-to-fine approach that addresses the unidirectional bias of generative autoregressive models and their exposure bias as well as the lacking global context. 
We formulate learning the data density as a hierarchical problem. A coarser stage provides compressed contextual side information about the \emph{entire} image for the autoregressive process on the next finer stage. 
We utilize a diffusion process to gradually eliminate information and compress the data, yielding a hierarchy of increasingly abstract and compact representations. 
The first scale of this approach is a discrete
representation learning task (cf.
\citep{DBLP:conf/nips/OordVK17,DBLP:conf/nips/RazaviOV19,DBLP:journals/corr/abs-2005-00341,DBLP:journals/corr/abs-2012-09841,DBLP:journals/corr/abs-2104-10157,DBLP:journals/corr/abs-2102-12092}). Subsequently, we further compress this learned representation via a fixed,
multinomial diffusion process \citep{DBLP:journals/corr/Sohl-DicksteinW15,DBLP:journals/corr/abs-2102-05379}. We then invert this process by training a Markov chain to recover the data from this hierarchy. Each Markovian transition is modeled autoregressively but it simultaneously attends to the preceding state in the hierarchy, which provides crucial global context to each individual autoregressive step. As each of this steps can also be interpreted as
learning a denoising cloze task \citep{DBLP:conf/acl/LewisLGGMLSZ20}, where
missing tokens at the next finer stage are ``refilled'' with a bidirectional encoder and an autoregressive decoder, we dub our approach \emph{ImageBART}.

\textbf{Contributions of our work}
Our approach tackles high-fidelity image synthesis
with autoregressive models by
learning to invert a fixed multinomial diffusion process 
in a 
discrete space of compact image representations to successively introduce context.
This reduces both the often encountered exposure bias of AR models and also 
enables locally controlled, user-interactive image editing.
Additionally, our model effectively handles a variety of conditional synthesis tasks and our introduced hierarchy corresponds to a successively compressed image representation. We observe that our model sample visually plausible images while still enabling a trade-off between reconstruction capability and compression rate.

\section{Related Work}
\enlargethispage{\baselineskip}
\textbf{Latent Variable Models}
Among likelihood-based approaches, latent variable models represent a data
distribution with the help of unobserved latent variables. 
For example, Variational Autoencoders (VAEs) \citep{VAE, VAE2} encode data points into a lower
dimensional latent variable with a factorized distribution. This makes them
easy to sample, interpolate \citep{DBLP:conf/iclr/LesniakSP19,
DBLP:journals/corr/abs-1710-11381} and modify
\citep{DBLP:journals/corr/White16a}. In a conditional setting
\citep{DBLP:journals/corr/KingmaRMW14},
latent variables which are independent from the conditioning lead to
disentangled representations \cite{DBLP:conf/eccv/JhaASV18,DBLP:conf/iclr/SzaboHPZF18,DBLP:journals/corr/MathieuZSRL16,DBLP:conf/nips/RombachEO20}.
A hierarchy of latent variables \citep{DBLP:conf/nips/SonderbyRMSW16} gives 
mutli-scale representations of the data. 
Unfortunately, even the %
deepest instantiations of these models
\citep{DBLP:conf/nips/MaaloeFLW19,DBLP:conf/nips/VahdatK20,DBLP:journals/corr/abs-2011-10650} lack in sample quality
compared to other generative models and are oftentimes restricted to highly
regular datasets. %

\textbf{Autoregressive Models}
AR models represent a distribution as a product of conditional, learnable
factors via the chain rule of probability densities. 
While this makes them powerful models for density estimation
\citep{DBLP:journals/corr/UriaCGML16, DBLP:journals/corr/GermainGML15}, their
samples often lack global consistency. Especially on image data modeled with
convolutional architectures \citep{DBLP:journals/corr/OordKVEGK16, DBLP:journals/corr/SalimansKCK17}, this has been attributed to a locality bias of
convolutional neural networks (CNNs) which biases the model towards strong
local correlations between neighboring pixels at the expense of a proper
modeling of coherence \citep{DBLP:conf/icml/KolesnikovL17,DBLP:journals/corr/abs-1903-04933}. This leads to samples resembling texture patterns without discernible global structure. 
Attempts to fix this properties by including explicit latent variables \citep{DBLP:journals/corr/GulrajaniKATVVC16, DBLP:journals/corr/ChenKSDDSSA16, DBLP:journals/corr/abs-1903-04933} have not been overly successful, mainly due the
expressiveness of AR models, providing little incentive for learning additional
latent variables.

\textbf{Generative Models on Improved Representations}
Another successful line of work first learn an improved image representation
and subsequently learn a generative model for this representation
\citep{DBLP:conf/nips/OordVK17,DBLP:conf/iclr/DaiW19}. Most works
\citep{DBLP:conf/nips/RazaviOV19, DBLP:journals/corr/abs-2012-09841, DBLP:journals/corr/abs-2102-12092}
learn a discrete representation which is subsequently modeled autoregressively
but approaches using continuous representations in combination with
VAEs \citep{DBLP:conf/iclr/DaiW19}, or normalizing flows \citep{10.1145/3447648, DBLP:conf/nips/RombachEO20, DBLP:journals/corr/abs-2004-13166}, exist too.
Learning a compact representation enables the use of
transformers for autoregressive modeling \cite{DBLP:conf/icml/ChenRC0JLS20}, which avoids the locality bias of
CNNs, can be used for the synthesis of complex scenes conditioned on text as in DALL-E \cite{DBLP:journals/corr/abs-2102-12092}, and, when combined with adversarial learning \citep{goodfellow2014GAN}, enables sampling of coherent
high-resolution images \citep{DBLP:journals/corr/abs-2012-09841}.
However, AR modeling of a learned representation still limits applications
compared to latent variable models. Their samples can still exert artifacts resulting from a
sequential modeling of components, and, since these models are always trained 
by ``teacher-forcing'', they are susceptible to an exposure bias
\citep{DBLP:journals/corr/BengioVJS15,DBLP:journals/corr/RanzatoCAZ15,DBLP:conf/nips/GoyalLZZCB16,DBLP:conf/emnlp/Schmidt19,DBLP:journals/corr/LeblondAOL17}.

\textbf{Diffusion Probabilistic Models}
Diffusion probabilistic models revert a fixed,
diffusion process with a learned Markov
Chain~\cite{DBLP:journals/corr/Sohl-DicksteinW15}. 
Being directly applied in pixel space,
however, downstream analysis reveals that these models tend to
optimize subtle details of the modeled data, which have little contribution to
the sample quality~\citep{DBLP:conf/nips/HoJA20, DBLP:journals/corr/abs-2105-05233}, particularly hindering applications on
high-resolution and -complexity datasets. By using a multinomial diffusion process \citep{DBLP:journals/corr/abs-2102-05379} (recently generalized by \citep{austin2021structured}) on
a compressed, discrete representation of images, we circumvent these
issues. Diffusion probabilistic models require a very large number of diffusion
steps in order to model the reverse process with a model distribution that
factorizes over components. Because our approach uses autoregressively
factorized models for the reverse process, we can reduce the required number of
steps and obtain significant improvements in sampling speed and the ability to
model complex datasets. 
\modelfigure

\section{Method}
\label{sec:method}
\newcommand*{\approxident}{%
  \mathrel{\vcenter{\offinterlineskip
  \hbox{$\sim$}\vskip-.35ex\hbox{$\sim$}\vskip-.35ex\hbox{$\sim$}}}}
\newcommand{\RR}{\mathbb{R}}
\newcommand{\xpixel}{x_\text{pixel}}
\newcommand{\xpixelrec}{\tilde{x}_\text{pixel}}
\newcommand{\hpixel}{H}
\newcommand{\wpixel}{W}
\newcommand{\cpixel}{3}
\newcommand{\xt}[1]{x_{#1}}
\newcommand{\enc}{E}
\newcommand{\dec}{D}
\newcommand{\ppixel}{p_\text{pixel}}
\newcommand{\pxt}[1]{p_{\xt{#1}}}
\newcommand{\qxt}[1]{q_{\xt{#1}}}
\newcommand{\q}{q}

\subsection{Hierarchical Generative Models}
\vspace{-2mm}
\newcommand{\LPIPS}{\text{LPIPS}}
\newcommand{\KL}{\mathbb{KL}}
\newcommand{\expect}{\mathbb{E}}
\newcommand{\pmodel}[1]{p^{#1}_{\theta}}
\newcommand{\pchain}{p_{\theta}}
\newcommand{\qchain}{q_{\theta}}
\newcommand{\qmodel}[1]{q^{#1}_{\theta}}

To tackle the difficult problem of modeling a highly complex distribution
$p(x)$ of high-dimensional images $x$, we (i) introduce bidirectional
context into an otherwise unidirectional autoregressive factorization of $p(x)$
as in Eq.~\eqref{eq:ar} and (ii) reduce the difficulty of the learning problem with a
hierarchical approach. To do so, we learn a sequence of distributions
$(\pmodel{t})_{t=0}^T$, such that each distribution $\pmodel{t-1}$ models a
slightly more complex distribution with the help of a slightly simpler
distribution $\pmodel{t}$ one level above.
This introduces a coarse-to-fine hierarchy of image representations $\xt{0:T}\coloneqq(\xt{t})_{t=0}^T$,
such that an $\xt{t-1}$ is modeled conditioned on $\xt{t}$, i.e. $\xt{t-1}
\sim \pmodel{t-1}(\xt{t-1} | \xt{t})$ and defines a reverse Markov Chain for
$x\eqqcolon\xt{0}$ as $\pchain(\xt{0}) = \pmodel{T}(\xt{T}) \prod_{t=1}^T
\pmodel{t-1}(\xt{t-1} | \xt{t})$.
Since our goal is to approximate the original distribution $p(x)$ with
$\pchain(\xt{0})$, we introduce a forward Markov Chain,
$\qchain(\xt{1:T}|\xt{0})=\prod_{t=1}^T \qmodel{t}(\xt{t}|\xt{t-1})$, to obtain
a tractable upper bound on the Kullback-Leibler (KL) divergence between $p$ and
$\pchain$, $\KL(p(\xt{0}) \Vert \pchain(\xt{0})) =: \mathcal{KL}$, using the evidence lower bound (ELBO). With
$\qmodel{T}(\xt{T}|\xt{T-1}) \coloneqq \pmodel{T}(\xt{T})$, we obtain
\begin{align}
 \label{eq:elbo}
    \MoveEqLeft
    \mathcal{KL} \leq %
    \underbrace{\expect_{\xt{0}, \xt{1}} \log
    \frac{p(\xt{0})}{\pmodel{0}(\xt{0} | \xt{1})}}_{\eqqcolon L_1 \rightarrow \text{ discrete repr. learning}} %
    + \sum_{t=2}^T \underbrace{\expect_{\xt{0}, \xt{t}} \KL\big(\qmodel{t-1}(\xt{t-1}|\xt{t},\xt{0})
    \Vert \pmodel{t-1}(\xt{t-1}|\xt{t})\big)}_{\eqqcolon L_t \rightarrow\text{ decoupled with diffusion process}} %
\end{align}

We use $L_1$ to learn a compressed and discrete representation of images, such
that subsequent stages of the hierarchy do not need to model redundant
information (Sec.~\ref{sec:vqgan}).
With $L_t, t>1$ we learn a model that can rely on global context from a
coarser representation $\xt{t}$ to model the representation $\xt{t-1}$
(Sec.~\ref{sec:cloze}). See Fig.~\ref{fig:modelfigure} for an overview of the proposed model.

\subsection{Learning a compact, discrete representation for images}
\label{sec:vqgan}
\enlargethispage{\baselineskip}
\vspace{-0.5em}
\newcommand{\seqlength}{N}
\newcommand{\codebooksize}{K}

\newcommand{\lrec}{L_{rec}}
\newcommand{\ladv}{L_{adv}}
\newcommand{\lcb}{L_{cb}}

Since the first stage of the hierarchical process is the one that operates directly on the data, we assign it a separate role. 
To avoid that the optimization of $L_t$ ($t=1,\dots,T$) in Eq.~\eqref{eq:elbo}
unnecessarily wastes capacity on redundant details in the input images---which is an often encountered property of pixel-based likelihood models
\citep{DBLP:conf/nips/OordVK17,DBLP:journals/corr/abs-2012-09841,DBLP:journals/corr/abs-2103-03841}---we take $L_1 = \expect_{p(\xt{0})\qmodel{1}(\xt{1}|\xt{0})} \log \frac{p(\xt{0})}{\pmodel{0}(\xt{0} | \xt{1})}$ to be the reconstruction term for a discrete autoencoder model. This has the advantage that we can directly build on work in neural discrete representation learning, which has impressively demonstrated that discrete representations can be used for high-quality synthesis of diverse images while achieving strong compression. In particular, \citep{DBLP:conf/nips/MentzerTTA20}
and \citep{DBLP:journals/corr/abs-2012-09841} have shown that adding an adversarial realism prior to the usual autoencoder objective
helps to produce more realistic images at higher compression rates by locally trading 
reconstruction fidelity for realism.

More specifically, we follow \citep{DBLP:journals/corr/abs-2012-09841} to encode images into a low-dimensional 
representation which is then vector-quantized with a learned codebook of size $\codebooksize$ to obtain
$\{0,\dots,\codebooksize-1\}^{h \times w} \ni \xt{1}~\sim
\qmodel{1}(\xt{1}|\xt{0})$ deterministically as the index of the closest
codebook entry.
The encoder is a convolutional neural network
(CNN) with four 
downsampling steps, such that $h=H/16$ and $w=W/16$ for 
any input image $x_0 \in \mathbb{R}^{H \times W \times 3}$. 
For downstream autoregressive learning, this representation is then unrolled into
a discrete sequence of length $\seqlength = h \cdot w$.
To recover an image from $\xt{1}$, we utilize a CNN decoder $G$, such that the 
reverse model is specified as
\begin{equation}
- \log \pmodel{0}(\xt{0} | \xt{1}) \propto f_{rec}(\xt{0}, G_\theta(\xt{1})) + \log D_\phi(G_\theta(\xt{1})) =: \lrec(x_0, x_1; \theta) + \ladv(x_1; \theta, \phi)
\label{eq:firststagep}
\end{equation}
Here, $f_{rec}$ denotes the perceptual similarity metric
\citep{DBLP:conf/cvpr/GatysEB16,DBLP:conf/eccv/JohnsonAF16,DBLP:conf/nips/DosovitskiyB16,lpips} (known as $\LPIPS$) and $D_\phi$
denotes a patch-based adversarial discriminator \citep{goodfellow2014GAN}. 
Note that, due to the deterministic training, the likelihood in Eq.~\eqref{eq:firststagep} is likely to be degenerate.
$D_\phi$ is optimized to differentiate original images $\xt{0}$ from their reconstruction $G_\theta(\xt{1})$
using simultaneous gradient ascent, such that the objective for learning 
the optimal parameters $\{\theta^*, \phi^*\}$ reads:
\begin{equation}
\{\theta^*, \phi^*\} = \arg \min_\theta \max_\phi \Big(\lrec(x_0, x_1; \theta) - \ladv(x_1; \theta, \phi) + \log D_\phi(x_0) + \lcb(\theta) \Big)
\label{eq:vqganobjective}
\end{equation}
The optimization of $\theta$ via this objective includes the parameters of the encoder and decoder in addition to the parameters of the learned codebook, trained via the codebook loss $\lcb$ as in \citep{DBLP:conf/nips/OordVK17, DBLP:journals/corr/abs-2012-09841}.

\subsection{Parallel learning of hierarchies}
\label{sec:cloze}
\vspace{-0.5em}
Under suitable choices for $\pchain, \qchain$, one can directly optimize these
chains over $\sum_t L_t$. However, the objectives $L_t$ of the hierarchy levels are coupled through the
forward chain $\qchain$, which makes this optimization problem difficult.  With
expressive reverse models $\pmodel{t-1}$, the latent variables $\xt{t}$ are often ignored by
the model \citep{DBLP:journals/corr/abs-1903-04933} and the scale of the different level-objectives can be
vastly different, resulting in a lot of gradient noise that hinders the
optimization \citep{DBLP:journals/corr/abs-2102-09672}. In the continuous case, reweighting schemes for
the objective can be derived \citep{DBLP:conf/nips/HoJA20} based on a connection to score
matching models \citep{DBLP:conf/nips/SongE19}. However, since we are working with a discrete
$\xt{1}$, there is no analogue available.

While we could follow the approach taken for the first level and
sequentially optimize over the objectives $L_t$,
this is a rather slow process since each level $t-1$
needs to be converged before we can start solving level $t$. However, this sequential
dependence is only introduced through the forward models $\qmodel{t}$ and since
$\qmodel{1}$ already learns a strong representation, we can choose simpler
and fixed, predefined forward processes for $\qmodel{t}, t>1$. 
The goal of these processes, i.e., generating a hierarchy of distributions by reducing 
information in each transition, 
can be readily achieved by, e.g., randomly masking \cite{DBLP:journals/corr/abs-1810-04805}, removing \cite{DBLP:conf/acl/LewisLGGMLSZ20} or replacing \cite{DBLP:journals/corr/abs-2102-05379} a fraction of the components of $\xt{t-1}$.

\paragraph{Multinomial diffusion}
\newcommand{\cat}{\mathcal{C}}
\newcommand{\one}{\mathbbm{1}}
\newcommand{\alphacum}{\bar{\alpha}}
This process of randomly replacing a fraction $\beta_t$ of the components with random
entries can be described as a multinomial diffusion process \citep{DBLP:journals/corr/abs-2102-05379}, a natural
generalization of binomial diffusion \citep{DBLP:journals/corr/Sohl-DicksteinW15}. The only parameter
$\theta$ of
$\qmodel{t}$ is therefore $\beta_t$, which we consider to be fixed.
Using the standard basis $e(k)=(\delta_{jk})_{j=1}^\codebooksize$,
the forward process can be written as a
product of categorical distributions $\cat$ specified in terms of the
probabilities over the codebook indices:
\begin{equation}
  \qmodel{t}(\xt{t} | \xt{t-1}) = \prod_{i=1}^\seqlength \cat(\xt{t}^i |
  (1-\beta_t)e(\xt{t-1}^i) + \beta_t \one/\codebooksize), \quad t>1
  \label{eq:mdiff}
\end{equation}
where $\one=(1)_{j=1}^\codebooksize$ is the all one vector. It then follows
that after $t-1$ steps, on average, a fraction of $\alphacum_t \coloneqq
\prod_{l=2}^t (1-\beta_t)$ entries from $\xt{1}$ remain unchanged in $\xt{t}$, i.e.
\begin{equation}
  \qmodel{t}(\xt{t}|\xt{1}) = \prod_{i=1}^\seqlength \cat(\xt{t}^i |
  \alphacum_t e(\xt{1}^i) + (1-\alphacum_t) \one/\codebooksize), \quad t>1.
\end{equation}
This enables computation of the posterior $\qchain(\xt{t-1}|\xt{t},\xt{1}) =
\frac{\qmodel{t}(\xt{t}|\xt{t-1})\qchain(\xt{t-1}|\xt{1})}{\qchain(\xt{t}|\xt{1})}$
for $t>2$,
and, using the fact that $\qmodel{1}$ is deterministic,
we can rewrite $L_t$ as
\begin{equation}
\expect_{p(\xt{0})}\expect_{\qchain(\xt{t}|\xt{1})} \KL\big(\qmodel{t-1}(\xt{t-1}|\xt{t},\xt{1})
    \Vert \pmodel{t-1}(\xt{t-1}|\xt{t})\big), \quad t>2
\label{eq:posteriormatch}
\end{equation}
such that the KL term can now be computed analytically for $t>2$. For $t=2$, we use
a single sample Monte-Carlo estimate for the maximum likelihood reformulation,
i.e.
\begin{equation}
  \argmin L_2 = \argmax
    \expect_{p(x_0)} \expect_{\qmodel{2}(\xt{2}|\xt{1})} \log
    \pmodel{1}(\xt{1}|\xt{2}).
\end{equation}
Finally, we set $\pmodel{T}$ to be a uniform distribution. This completes the
definition of the reverse chain $\pchain$, which can now be started from a random
sample for $\xt{T}\sim\pmodel{T}(\xt{T})$, denoised sequentially through
$\xt{t-1}\sim\pmodel{t-1}(\xt{t-1}|\xt{t})$ for $t=T,\dots,2$, and finally be
decoded to a data sample $\xt{0}=G(\xt{1})$.

\paragraph{Reverse diffusion models}
Under what conditions can we recover the true data distribution? By rewriting
$\sum_t L_t$, we can see from
\begin{equation}
  \KL(p(\xt{0}) \Vert \pchain(\xt{0})) \leq \sum_{t=1}^T
  \KL(\qchain(\xt{t-1}|\xt{t}) \Vert \pmodel{t-1}(\xt{t-1}|\xt{t}))
\end{equation}
that this is possible
as long as all reverse models are expressive enough to represent the true reverse
processes defined by $\qchain$.
For the first level, we can ensure this by making $\xt{1}$
large enough such that the reconstruction error becomes negligible.
For the diffusion process, previous image models \cite{DBLP:journals/corr/Sohl-DicksteinW15,DBLP:conf/nips/HoJA20,DBLP:journals/corr/abs-2011-13456,DBLP:journals/corr/abs-2102-05379}
relied on the fact that, in the limit $\beta_t \to 0$, the form of the true
reverse process has the same functional form as the forward diffusion process
\cite{DBLP:journals/corr/Sohl-DicksteinW15,Kolmogoroff1931}. In particular, this allows modeling of the
reverse process with a distribution factorized over the components. However, to
make $\qmodel{T-1}$ close to a uniform distribution
requires a very large $T$ (in the order of 1000 steps) with small $\beta_t$. Training such a large
number of reverse models is only feasible with shared weights for the models,
but this requires a delicate reweighting \cite{DBLP:conf/nips/HoJA20} of the objective and currently no
suitable reweighting is known for the discrete case considered here.

Thus, to be able to recover the true data distribution with a modest number of
reverse models that can be trained fully parallel, and without weight-sharing, we
model each reverse process autoregressively. We use an encoder-decoder
transformer architecture \cite{DBLP:conf/nips/VaswaniSPUJGKP17}, such that the
decoder models the reverse process for $\xt{t-1}$ autoregressively with the help of global
context obtained by cross-attending to the encoder's representation of
$\xt{t}$ as visualized in Fig.~\ref{fig:modelfigure}. Note that the need for autoregressive modeling gets reduced for small
$\beta_t$, which we can adjust for by reducing the number of decoder
layers compared to encoder layers. The use of the compression model described in
Sec.~\ref{sec:vqgan}, however, allows to utilize full-attention based transformer architectures to implement the autoregressive scales. 
\modelsamplessmall
\section{Experiments}
\vspace{-2mm}
Sec.~\ref{subsec:expone} evaluates the quality ImageBART achieves in image synthesis.
Since we especially want to increase the controllability of the generative
process, we evaluate the performance of ImageBART on class- and text-conditional
image generation in Sec.~\ref{subsec:exptwo}. The ability of our approach to attend
to global context enables a new level of localized control which is not
possible with previous, purely autoregressive approaches as demonstrated in
Sec.~\ref{subsec:expthree}. Finally, Sec.~\ref{subsec:expfour} presents
ablations on model and architecture choices.
\vspace{-1em}
\subsection{High-Fidelity Image Synthesis with ImageBART} %
\label{subsec:expone}
\vspace{-0.8em}
In this section we present qualitative and quantitative results on images
synthesized by our approach. We train models at resolution $256\times 256$ for unconditional generation on
FFHQ \citep{stylegan}, LSUN -Cats, -Churches and -Bedrooms \citep{DBLP:journals/corr/YuZSSX15}
and on class-conditional synthesis on ImageNet (cIN) \citep{DBLP:conf/cvpr/DengDSLL009}.
\fids
\quantcond

\textbf{Effective Discrete Representations}
Learning the full hierarchy
as described in Eq.~\eqref{eq:elbo} and without
unnecessary redundancies in the data
requires to first learn a strong compression model via
the objective in Eq.~\eqref{eq:vqganobjective}.
\citep{DBLP:journals/corr/abs-2012-09841} demonstrated how to effectively
train such a model and we directly utilize the
publicly available pretrained models.
For training on LSUN, we finetune an ImageNet pretrained model for one epoch
on each dataset.
As the majority of codebook entries remains unused,
we shrink the codebook to those entries which are actually used
(evaluated on the validation split of ImageNet) and assign a
random entry for eventual outliers.
This procedure yields an effective, compact
representation on which we subsequently train ImageBART.

\textbf{Training Details}
As described in Sec.~\ref{sec:cloze}, we use
an encoder-decoder structure
to model the reverse Markov Chain $\pmodel{t-1}(\xt{t-1}|\xt{t}), \; t < T$,
where the encoder is a bidirectional transformer model
and decoder is implemented as an AR transformer.
As the context for the last scale is pure noise,
we employ a decoder-only variant to model $\pmodel{T-1}(\xt{T-1}|\xt{T})$.
Furthermore, to account for the different complexities of the datasets,
we adjust the number of multinomial diffusion steps
for each dataset accordingly. %
For FFHQ we choose a chain of length
$T = 3$, such that the total model consists of (i) the
compression stage and (ii) $n=2$
transformer models trained in parallel via the objective
described in Eq.\eqref{eq:posteriormatch}. Similarly, we set $n=3$
for each of the LSUN models and $n=5$ for the ImageNet model.

\textbf{Results}
For each of these settings, Fig.~\ref{fig:modelsamplessmall} depicts samples of size $256\times256$ generated
with ImageBART and a single pass through the learned Markov Chain, demonstrating that our model is able
to produce realistic and coherent samples. This is further confirmed by a quantitative analysis in Tab.~\ref{tab:fids}, where we compare FID scores of
competing likelihood-based and score-based methods such as TT \citep{DBLP:journals/corr/abs-2012-09841} and DDPM \citep{DBLP:conf/nips/HoJA20}.
Regarding other works on diffusion models such as \citep{DBLP:conf/nips/HoJA20} and \citep{DBLP:journals/corr/abs-2011-13456} operating directly in pixel space, we observe that these approaches perform roughly equivalently well in terms of FID
for datasets of low complexity (e.g. LSUN-Bedrooms and-Churches). For more complex datasets (LSUN-Cats, cIN),
however, our method outperforms these pixel-based approaches, which can also be
seen qualitatively on the right in Tab.~\ref{tab:fids}.
See Fig.~\ref{fig:cincomp} for a comparison on ImageNet.

\subsection{Conditional Markov Chains for Controlled Image Synthesis}
\label{subsec:exptwo}
\vspace{-2mm}
Being a sequence-to-sequence model, our approach allows for flexible and
arbitrary conditioning by simply preprending tokens, similar to \citep{DBLP:journals/corr/abs-2012-09841, DBLP:journals/corr/abs-2102-12092}.
More specifically, each learned transition $\pmodel{t-1}(\xt{t-1}|\xt{t}, c), \: t > 1$
of the Markov chain is then additionally conditioned on a representation $c$,
e.g. a single token in the case of the class-conditional model of Sec.~\ref{subsec:expone}.
Note that the compression model $\pmodel{0}$ remains unchanged.

\textbf{Text-to-Image Synthesis}
\conceptualcaptionsamples
Besides class-conditional modeling on ImageNet, we also learn a text-conditional
model on \emph{Conceptual Captions} (CC) \citep{sharma2018conceptual, ng2020understanding}.
We obtain $c$ by using the publicly available tokenizer of the CLIP model \citep{DBLP:journals/corr/abs-2103-00020},
yielding a conditioning sequence of length 77. To model the dataset, we choose $T=5$ and thus
train $n=4$ transformer models independently. For the $\pmodel{0}$, we directly transfer the
compression model from Sec.~\ref{subsec:expone}, trained on the ImageNet dataset.

Fig.~\ref{fig:conceptualcaptionsamples} visualizes
 synthetic samples obtained with this model for various ``image-cloze''
tasks. Our resulting model is able to attend to semantic variations in the
conditioning sentence (e.g. a change of weather for imagery of mountains)
and renders the corresponding images accordingly.
In Tab.~\ref{fig:quant_cond1}, we evaluate FID \cite{FID} and Inception Scores
(IS) \cite{Salimans2016ImprovedTF} to measure the quality of synthesized
images, as well as cosine similarity between CLIP
\cite{DBLP:journals/corr/abs-2103-00020} embeddings of the text prompts and the
synthesized images to measure how well the image reflects the text.
ImageBART improves all metrics upon \cite{DBLP:journals/corr/abs-2012-09841}. Fig.~\ref{fig:comptext2imgsupp} in the supplement provides corresponding qualitative examples for user-defined text inputs.

\textbf{Resolutions Beyond $\boldsymbol{256 \times 256}$ Pixels.}
Our approach is not restricted to generating images of size $256 \times 256$ pixels.
Although trained on a fixed resolution, we can apply our models in a patch-wise manner,
where we use the sliding attention window of \citep{DBLP:journals/corr/abs-2012-09841} for each scale $t > 0$.
As we now incorporate more and more global context while decoding with
the Markov chain (which can be thought of as widening a noisy
receptive field), ImageBART is able to render consistent
images in the megapixel regime. See for example Fig.~\ref{fig:landscapeinterpolator},
where we use our text-conditional model to render an image of size
$300\times 1800$ pixel and interpolate between two different
text prompts.
More examples, especially also for semantically guided synthesis, can be found in 
Sec.~\ref{subsec:morecond}.
\landscapeinterpolator

\subsection{Beyond Conditional Models: Local Editing with Autoregressive Models}
\label{subsec:expthree}
\vspace{-2mm}
Recent autoregressive approaches, which use a CNN to learn a discrete
representation \citep{DBLP:conf/nips/OordVK17}, partially alleviate the issues of pixel-wise autoregressive models
by working on larger image patches. However, as we show in Fig.~\ref{fig:arcomp}, even approaches which use
adversarial learning to maximize the amount of context encoded in the discrete
representation \citep{DBLP:journals/corr/abs-2012-09841} cannot produce completions of the upper half of
an image which are consistent with a given lower half.

While our approach also models each transition autoregressively from the
top-left to the bottom-right, the ability to attend to global context from the
previous scale enables consistent completions of arbitrary order, e.g. right-to-left.
To achieve this, we mask the diffusion process as described in Sec.~\ref{supp:maskeddenoising}. 
For a user-specified mask $m$ (e.g.
the upper half of an image as in Fig.~\ref{fig:arcomp}), this results
in a forward-backward process $\pmodel{t-1|t-1,m}$, which,
by definition, leaves the unmasked context intact.
The reverse process then denoises the unmasked entries to make them consistent with
the given context. 

Fig.~\ref{fig:arcomp} (bottom) visualizes this mixing process, where we use a model with $T=3$. The first column shows the masked
input. To start the process we set all masked entries to random entries. The
first two columns then show (decoded) samples from the masked reverse processes $\pmodel{2,m}$ and
$\pmodel{1,m}$, which still display inconsistencies. The remaining columns show
the trajectory of the process $\pmodel{1|1,m}$, which demonstrates how the
model iteratively adjusts its samples according to the given context until it
converges to a globally consistent sample.
For illustration, we show the analog trajectory obtained with \citep{DBLP:journals/corr/abs-2012-09841},
but because it can only attend to unidirectional context, this trajectory is
equivalent to a sequence of independent samples and therefore fails to achieve
global consistency.

The masked process can be used with arbitrary masks, which enables localized
image editing with free, hand-drawn masks as shown in Fig.~\ref{fig:localediting}. Note that
our model does not need to be trained specifically for this task, which also
avoids generalization problems associated with training on masks
\citep{DBLP:journals/corr/abs-2104-00845}. 
Combining this property with the conditional
models from Sec.~\ref{subsec:exptwo} allows for especially interesting novel applications, where
local image regions are modified based on user specified class or text prompts,
as shown in Fig.~\ref{fig:localcondediting}.

\comparisonar
\ffhqinpainting
\conditionalinpainting
\subsection{Ablations}
\label{subsec:expfour}
\textbf{On the Number of Diffusion Steps}
In this section we analyze the effect of varying the number of diffusion steps
(denoted by $T$). To do so, we perform an experiment for unconditional training
on the FFHQ dataset, where we train a Taming Transformers (TT) baseline
(corresponding to the case $T=2$ within our framework) with 800M parameters and
three variants of ImageBART with $T=3$ (2x400M), $T=5$ (4x200M) and $T=9$
(8x100M), respectively.
Note that for a fair comparison, all models use the same first level for
compression, and we fix the number of remaining parameters to 800M and
distribute them equally across all scales. All models were trained with the
same computational budget and evaluated at the best validation checkpoint.

In Tab.~\ref{tab:tablation}, we assess both the pure synthesis and the modification
ability of ImageBART by computing FID scores on samples and modified images (in
the case of upper half completion as in Fig.~\ref{fig:arcomp}). For both tasks,
we use a single pass through the reverse Markov chain.
We observe that the modification performance increases monotonically with the
number of scales, which highlights the improved image manipulation abilities of
our approach. For unconditional generation, we observe a similar trend, although
FID seems to plateau beyond $T=5$.

\tablation

\textbf{Joint vs. Independent Training}
While it is possible to optimize Eq.~\eqref{eq:elbo} jointly 
across all scales, we found that training is more robust when
training all scales independently. Besides the usual separation of 
training the compression model $\pmodel{0}$ and the generative
model  $\pmodel{t\geq1}$, training the latter in parallel over multiple scales avoids the
tedious weighting of the loss contribution from different scales; an often encountered 
problem in other denoising diffusion probabilistic models \citep{DBLP:conf/nips/HoJA20}.%

\textbf{Efficiency with Less Decoder Layers}
As we implement the conditional transition probabilities $\pmodel{t-1}$ with an encoder-decoder transformer architecture, we are interested
in the effect of altering the ratio of encoder and decoder layers in the model.
Recent work has provided evidence that it is possible to significantly 
reduce the number of decoder layers and thus also decrease autoregressive decoding speed
while maintaining high quality \citep{kasai2021deep}.
\encoderdecoderablation
We perform an experiment on LSUN-Churches, where we analyze the 
effect of different layer-ratios on synthesis quality (measured by FID) 
and on decoding speed when fixing the total number of model parameters to 200M. 
The results in the left part of Fig.~\ref{fig:ablation} confirms that it is indeed
possible to reduce the number of decoder layers while maintaining satisfactory FID scores
with higher decoding efficiency. We identity a favorable trade-off between four and six 
decoder layers and transfer this setting to our other experiments.

Finally, we compare our model in terms of sampling speed with the recent state-of-the-art generative diffusion~\citep{DBLP:conf/nips/HoJA20, DBLP:journals/corr/abs-2011-13456} and AR models~\citep{DBLP:journals/corr/abs-2012-09841}. The results are summarized in Fig.~\ref{fig:ablation}.
While consistently being faster than all pixel-based models due to training in a compressed latent space, 
the increase in runtime w.r.t. \citep{DBLP:journals/corr/abs-2012-09841} is moderate due to the use of encoder-decoder transformers, i.e., a a decrease in pure decoder layers. If a faster runtime is desired, the speed can be further increased by reducing the number of decoder layers even more, see also the discussion in Sec.~\ref{suppsec:samplingspeed}.

\section{Conclusion}
\vspace{-0.5em}
\enlargethispage{\baselineskip}
We have proposed ImageBART, a hierarchical approach to introduce bidirectional
context into autoregressive transformer models for high-fidelity controllable
image synthesis. We invert a multinomial diffusion process by training a Markov
chain to gradually incorporate context in a coarse-to-fine manner. Our study
shows that this approach (i) introduces a natural hierarchical representation
of images, with consecutive levels carrying more information than previous
ones. (see also Fig.~\ref{fig:compressionfigure}). 
(ii) It alleviates the unnatural unidirectional ordering of pure
autoregressive models for image representation through global context from
previous levels of the hierarchy. (iii) It enables global and local
manipulation of a given input, a feat previously out-of-reach for ARMs. (iv) We
additionally show that our model can be efficiently conditioned on various
representations, allowing for a large class of conditional image synthesis
tasks such as semantically guided generation or text-to-image synthesis.

\FloatBarrier
\newpage

\appendix
\section{Appendix}
\subsection{Hyperparameters \& Implementation Details}
\FloatBarrier
\label{suppsec:hyper}

\subsubsection{Compression Models}
\compressionhyper
We follow \citep{DBLP:journals/corr/abs-2012-09841} and implement our image compression models as ``VQGANs''. More specifically, we
use the official implementation provided at \url{https://github.com/CompVis/taming-transformers} and fine-tune
the publicly available model for experiments on LSUN. For FFHQ, we train such a compression model from scratch.
See Tab.~\ref{tab:compressionhyper} for an overview. 
As some of the codebook entries remain unused after training, 
we shrink the codebook to its \emph{effective} size when training a generative model on top of it. For eventual entries not
detected during evaluation on the subset, we assign a random entry.

\subsubsection{Hierarchical Representations via Multinomial Diffusion}
\diffusionhyper
Tab.~\ref{tab:diffusionhyper} lists the configurations of the multinomial
diffusion processes for each experiment described in this work (see also
Tab.~\ref{tab:compressionhyper}).
Note that all representations $x_t$ for $T>1$ have the same spatial resolution,
but since each forward diffusion process gradually removes information, we
obtain a coarse-to-fine hierarchy. On average, level $x_t$ will contain 
$\lfloor \bar{\alpha}_t \cdot N \rfloor$ valid entries, which we denote as the
effective sequence length in Tab.~\ref{tab:diffusionhyper}.
Thus, ImageBART can also be interpreted as a
generative compression model as illustrated in Fig.~\ref{fig:compressionfigure}: By trading perfect reconstruction
quality for compression, one can obtain a significantly shorter sequence, still
representing a visually plausible image.  This provides the basis for learning
a generative model that does not waste capacity on redundancies in the data
\citep{dieleman2020typicality} and the compressed space significantly lowers
the computational demands for training and decoding.

\compressionfigure

\subsubsection{Reverse Diffusion with Transformer Models}
\transformerhyper
ImageBART is a learned Markov chain, trained to reverse the multinomial diffusion process described in Eq.~\eqref{eq:mdiff}.
We can efficiently model the conditionals $\pmodel{t}$ with a
sequence-to-sequence model and follow \citep{DBLP:conf/nips/VaswaniSPUJGKP17,
DBLP:conf/acl/LewisLGGMLSZ20} to implement $\pmodel{t}$ with an encoder-decoder
architecture. Tab.~\ref{tab:transformerhyper} summarizes the hyperparameters
used to implement the conditionals for each experiment. For comparison, the
models in Tab.~\ref{tab:fids} contain 115M (VDVAE), 255M (DDPM), 30M
(StyleGAN2), 158M (BigGAN), 448M (DCT) and 600M (TT) parameters.

\subsubsection{Hardware}
All models listed in Tab.~\ref{tab:compressionhyper} and Tab.~\ref{tab:transformerhyper} were optimized on a single NVIDIA A100 GPU and using 32-bit precision. Sampling speed as reported in Fig.~\ref{fig:ablation} was also measured on a NVIDIA A100.

\subsection{Details on Conditional Experiments}
\label{subsec:morecond}
\imagenetsamplessupp
\conceptualcaptionssupp

\textbf{Semantically Guided Synthesis}
In addition to class- and text-conditional generative modeling, we apply our
model to semantically guided synthesis of landscape images \citep{spade}.
To achieve this we follow \citep{DBLP:journals/corr/abs-2012-09841} 
and use the discrete representation of an autoencoder model 
trained on segmentation masks as conditioning $c$ for our models $\pmodel{t}$. 
However, since simply prepending $c$ here doubles the total length, 
which means a fourfold increase in complexity in the attention mechanism, 
we exploit the fact that the segmentation masks and the images (or their representations) are aligned. 
More specifically, within the encoder-decoder architecture, we first 
produce two embeddings $e_1$ and $e_2$ for $x_t$ and $c$, respectively, 
which are subsequently concatenated channel-wise,
thereby keeping the sequence length of $x_t$.
With this modifications, we train a model with $T=5$ and individually optimize each scale similar to the unconditional training setting. Here again, we use the compression model $\pmodel{0}$   
pre-trained on ImageNet. For training, we randomly crop the images and semantic maps to size $256 \times 256$. For testing, however, we again use the sliding window approach of~\citep{DBLP:journals/corr/abs-2012-09841} (cf. Sec.~\ref{subsec:exptwo}), which enables us to generate high-resolution images of landscapes, as visualized in  Fig.~\ref{fig:sflckronepage} and Fig.~\ref{fig:sflickrsamplesone}.

\sflickronepage
\sflckrconditionallarge

\subsection{Masked Diffusion Processes for Local Editing}
\label{supp:maskeddenoising}
\suppcomparisonar
\conditionalinpaintingsupp
\conditionalinpaintingsupptxt
\suppfoxchain

Previous autoregressive approaches \citep{NIPS2016_b1301141} model images
directly as a sequence of pixels from the top-left to the bottom-right. Thus,
when generating a pixel, only context from neighbors to the left and above can
be taken into account.  While more recent approaches, which use a CNN to learn
a discrete representation that is subsequently modeled autoregressively
\citep{DBLP:conf/nips/OordVK17,DBLP:journals/corr/abs-2012-09841}, improve this
situation because elements of the representation now correspond to image
patches, Fig.~\ref{fig:arcomp} showed that these models still fail to generate
completions of the upper half of an image which are consistent with a given
lower half.

While our approach also models each transition autoregressively from the
top-left to the bottom-right, each transition additionally has access to global
context from the previous step.
We aim to exploit this fact to obtain novel applications such as consistent
completions of upper halfs and, more generally, completions with respect to an
arbitrary mask. For any such mask, let $m$ denote the result of downsampling it
to the size of $\xt{1}$ using nearest-neighbor-interpolation,
such that $m^i=0$ gives the positions where context should be used, and $m^i=1$
gives the positions where new content should be generated. We then define the
masked forward process,
\begin{equation}
  \qmodel{t,m}(\xt{t}|\xt{t-1}) = m\cdot\qmodel{t}(\xt{t}|\xt{t-1}) +
  (1-m)\cdot\delta(\xt{t}-\xt{t-1}),
\end{equation}
which only diffuses masked entries, and the masked reverse process,
\begin{equation}
  \pmodel{t-1,m}(\xt{t-1}|\xt{t}) = m\cdot\pmodel{t-1}(\xt{t-1}|\xt{t}) +
  (1-m)\cdot\delta(\xt{t-1}-\xt{t}),
\end{equation}
which only denoises masked entries. By definition, running this process forward
and then backward again represents the identity on umasked entries such that
the given context remains constant. We denote this forward-backward process
that starts from a given $\xt{t-1}$ and produces a sample $\xt{t-1,m}$,
\begin{equation}
  \xt{t} \sim \qmodel{t,m}(\xt{t}|\xt{t-1}),\quad
  \xt{t-1,m} \sim \pmodel{t-1,m}(\xt{t-1,m}|\xt{t})
\end{equation}
by $\pmodel{t-1|t-1,m}$ and use it to sample with spatial
conditioning information. Since it always leaves the unmasked context intact,
the reverse process denoises the unmasked entries to make them consistent with
the given context.

Besides Fig.~\ref{fig:arcomp},~\ref{fig:localediting} additional visualizations of this process can be
found in Fig.~\ref{fig:supparcomp}. The top shows masked inputs (left), final
results of upper completions obtained by
\citep{DBLP:journals/corr/abs-2012-09841} (middle) and by $\pmodel{1|1,m}$
(right). The bottom visualizes the trajectory of the masked process, showing
the masked input (leftmost column), denoised samples from $\pmodel{2,m}$ (first
column) and $\pmodel{1,m}$ (second column), and every other sample from the
forward-backward model $\pmodel{1|1,m}$. It demonstrates how the model
iteratively incorporates global context from the previous scale to converge to
a globally consistent sample at the very right.
A visualization of the process
on the class conditional ImageNet model is shown in
Fig.~\ref{fig:suppfoxchain}.
Additional examples for
conditional samples from this process, as in Fig.~\ref{fig:localcondediting}, can be found in
Fig.~\ref{fig:supplocalcondediting} and Fig.~\ref{fig:supplocalcondeditingtext}.

\newpage
\subsection{Limitations and Societal Impacts}
Training deep generative models consumes a significant amount of energy (see
also Sec.~\ref{suppsec:hyper} regarding the used hardware; the ImageNet model
for example was trained for 19 days). With regard to the environment,
it is important that we reduce the energy consumption as much as possible. To
take a step in this direction, we followed previous works and relied on a
strongly compressed, learned representation of images. Because we can fine-tune
the corresponding encoder and decoder models from pre-trained ones, the costs
for this step are largely amortized and subsequent levels of our hierarchy
benefit from a drastically reduced sequence length. Nonetheless, it should be
noted that such a strong compression scheme for images does not result in
perfect reconstructions. For applications which require very high fidelity,
such a level of compression might be unsuitable due to artifacts in the
reconstructed images. Additionally, the use of adversarial learning in this
stage can potentiate biases of datasets by its mode-seeking behavior.
Both of these issues can be lessened with larger sequence lengths at the cost
of higher energy requirements.

The transformer architecture which is used to model the transitions in our
hierarchy is generally considered to be less biased compared to convolutional
architectures. However, it also cannot benefit from useful inductive biases and
therefore requires a large amount of data and resources to learn all required
relationships. In early experiments, we noticed that on small datasets, such as
CIFAR-10 \citep{Krizhevsky2009LearningML}, the transformer models overfit before they reach good
performance. Thus, in its current implementation our approach requires datasets
of sufficient size. Future works should evaluate different architectures,
regularizations or augmentations to enable its use on small datasets. On the
other extreme, we find that with large datasets, the main bottleneck is the
computational resources that can be spent on the training of the transformer
models. On the largest datasets, \emph{Conceptual Captions} and
\emph{ImageNet}, we find that performance still improves after two weeks of
training. Thus, consistent with other works on scaling up generative models, we
expect that performance of our model will keep increasing with the available
resources.

To ensure comparability with other approaches, we use standard
benchmark datasets for deep generative models, even if some of them are known
to contain offensive content \citep{crawford2019excavating}.

\subsection{Sampling Speed}
\label{suppsec:samplingspeed}
Here, we discuss the effects of varying the number of encoder vs. decoder layers in ImageBART on sampling speed as presented in Sec.~\ref{subsec:expfour}. On each diffusion scale, the encoder layers only have to run once whereas the decoder layers have to run $n_{\text{data\_dim}}$ times. This results in an approximate complexity of order $n_{\text{scales}} C(n_{\text{encoder\_layers}}+ n_{\text{data\_dim}} n_{\text{decoder\_layers}})$, where $C$ is the complexity of a single transformer layer. 
The speedup from such an encoder-decoder transformer over a decoder only transformer with $n_{\text{encoder\_layers}}+n_{\text{decoder\_layers}}$ layers is therefore 
\begin{equation}
\Big(n_{\text{encoder\_layers}}+n_{\text{decoder\_layers}}\Big) \cdot \Big(\frac{n_{\text{encoder\_layers}}}{n_{\text{data\_dim}}}+n_{\text{decoder\_layers}}\Big)^{-1}.
\label{eq:speedy}
\end{equation}

\subsection{Additional Samples \& Nearest Neighbors}
We provide additional samples from our models in Fig.~\ref{fig:cinone}-\ref{fig:ffhqsamplessupp}.
Additionally, we also provide nearest neighbors (measured in VGG feature space)
for our FFHQ and LSUN-Churches models in Fig.~\ref{fig:nnsffhq} and Fig.~\ref{fig:nnschurch}, respectively.

\cinone
\cintwo
\cincomp
\texttoimgcomptt
\churchessamplessupp
\catsgridsupp
\bedroomsgridsupp
\ffhqsamplessupp

\nnffhq
\nnchurches

\FloatBarrier
\newpage
{\small
\bibliographystyle{abbrvnat}
\bibliography{ms}
}

\newpage
\end{document}

%% file: figures.tex
\providecommand{\imwidth}{}

\providecommand{\impath}[1]{}
\providecommand{\impatha}[1]{}
\providecommand{\impathb}[1]{}
\providecommand{\impathc}[1]{}
\providecommand{\impathd}[1]{}
\providecommand{\impathe}[1]{}

\newcommand{\encoderdecoderablation}{
\begin{figure}%
\begin{minipage}[t]{.49\textwidth}
\includegraphics[width=\textwidth]{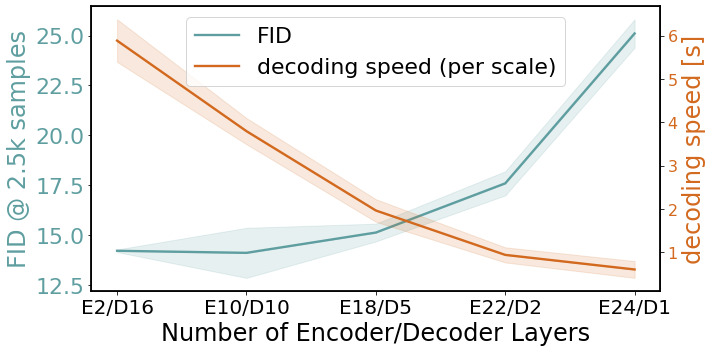}
\vfill
\end{minipage}
\hfill
\begin{minipage}[t]{.49\textwidth}
\includegraphics[width=\textwidth]{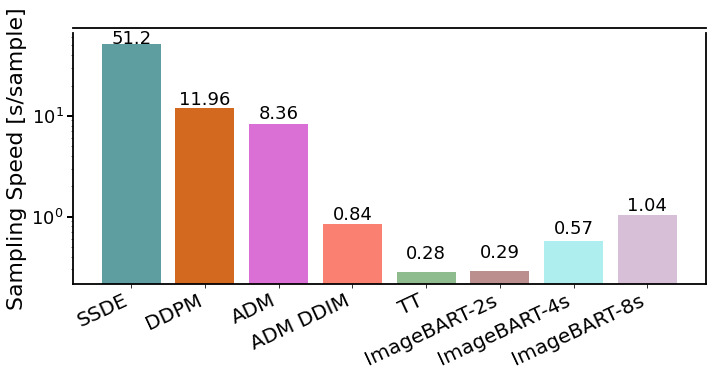}
\end{minipage}
	\caption{\label{fig:ablation} \emph{Left:} Effect of number of encoder vs. decoder layers for a fixed total number of model parameters ($(195 \pm 5) M$), evaluated on LSUN-Churches. FIDs are evaluated w.r.t 3 $\times$ 2500k samples. The plot shows 3 standard deviations. All models are trained jointly over three scales. \emph{Right:} Our model achieves better sampling performance than state of the art diffusion models (SSDE~\citep{DBLP:journals/corr/abs-2011-13456}, DDPM~\citep{DBLP:conf/nips/HoJA20}, ADM~\citep{DBLP:journals/corr/abs-2105-05233}) and also approaches the inference speed of TT~\citep{DBLP:journals/corr/abs-2012-09841}, which only consists of a single autoregressive stage. Reducing the number of scales increases inference speed at the expense of controllability. Experiments were conducted on a single NVIDIA A100 and are reported averaged over 1000 samples with a batch size of 50, evaluated on FFHQ while using the same number of trainable parameters (800m) for all AR models.
	\vspace{-1em}
	}
\end{figure}
}

\newcommand{\modelfigure}{
\begin{figure}[t]
	\centering
	\includegraphics[width=\textwidth]{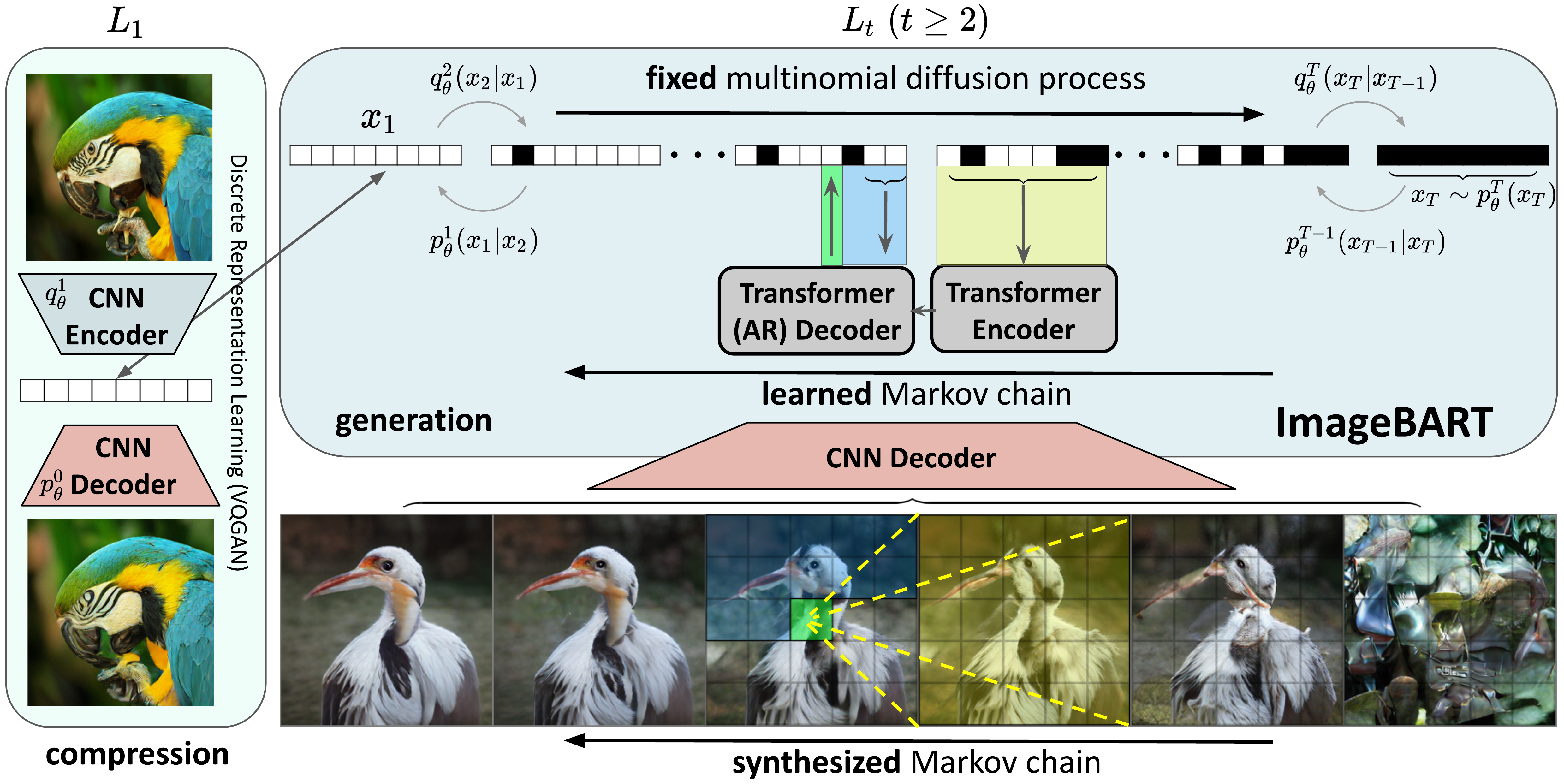}
\caption{\label{fig:modelfigure} Overview over our approach: We first learn a compressed, discrete image representation $x_1$ and subsequently our generative ImageBART model reverts a fixed multinomial diffusion process
        via a Markov Chain,
where the individual transition probabilities are modeled as independent autoregressive encoder-decoder models. This introduces a coarse-to-fine hierarchy such that each individual AR model can attend to global
context from its preceding scale in the hierarchy. \vspace{-1.5em}}
\end{figure}
}

\newcommand{\landscapeinterpolator}{
\begin{figure}%
	\centering
		\includegraphics[width=\textwidth]{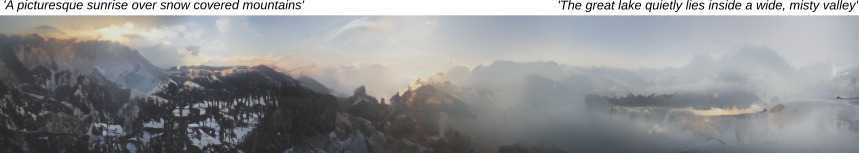}
	\caption{\label{fig:landscapeinterpolator}ImageBART is capable of generating high-resolution images. Here, we condition it on text prompts and interpolate between the two descriptions depicted above the image (see also Sec.~\ref{subsec:exptwo}).
	\vspace{-2em}}
\end{figure}
}

\newcommand{\compressionfigure}{
\begin{figure*}%
	\centering
    \renewcommand{\imwidth}{0.1\textwidth}
    \renewcommand{\impatha}[1]{img/compression/mushroom/400x267/##1} 
    \renewcommand{\impathb}[1]{img/compression/cat/256x256/##1} 
    \setlength{\tabcolsep}{1pt}
	\begin{scriptsize}
	\begin{adjustbox}{max width=\linewidth}
 	\begin{tabular}{c c cc@{\hskip 1em}ccc@{\hskip 1em}ccc}

	$x_0: r_b=1$ & $x_1: r_b=619$ & \multicolumn{2}{c}{$x_2: r_b=680$} & \multicolumn{3}{c}{$x_4: r_b=1190$} & \multicolumn{3}{c}{$x_5: r_b = 1.5 \cdot 10^5$}\\
  	\toprule
	\includegraphics[width=\imwidth]{\impatha{input}} &	         
	\includegraphics[width=\imwidth]{\impatha{vqrec}} &	         
	
	\includegraphics[width=\imwidth]{\impatha{scale1/0001}} &	         
	\includegraphics[width=\imwidth]{\impatha{scale1/0002}} &	         
	
	\includegraphics[width=\imwidth]{\impatha{scale4/0001}} &	         
	\includegraphics[width=\imwidth]{\impatha{scale4/0004}} &	         
	\includegraphics[width=\imwidth]{\impatha{scale4/0003}} &	         

	\includegraphics[width=\imwidth]{\impatha{scale5/0002}} &	         
	\includegraphics[width=\imwidth]{\impatha{scale5/0005}} &	         
	\includegraphics[width=\imwidth]{\impatha{scale5/0004}} \\

  	\midrule
	\includegraphics[width=\imwidth]{\impathb{0053_input}} &	         
	\includegraphics[width=\imwidth]{\impathb{0053_vqrec}} &	         
	
	\includegraphics[width=\imwidth]{\impathb{scale1/0053_0001}} &	         
	\includegraphics[width=\imwidth]{\impathb{scale1/0053_0006}} &	         
	
	\includegraphics[width=\imwidth]{\impathb{scale2/0053_0001}} &	         
	\includegraphics[width=\imwidth]{\impathb{scale2/0053_0006}} &	         
	\includegraphics[width=\imwidth]{\impathb{scale2/0053_0008}} &	         

	\includegraphics[width=\imwidth]{\impathb{scale3/0053_0040}} &	         
	\includegraphics[width=\imwidth]{\impathb{scale3/0054_0040}} &	         
	\includegraphics[width=\imwidth]{\impathb{scale3/0055_0034}} \\

	\end{tabular}
	\end{adjustbox}
	\end{scriptsize}
	\caption{\label{fig:compressionfigure} ImageBART can be interpreted as a generative compression model retaining high visual quality at high compression rates. Here, we denote the compression rate $r_b = \frac{256\cdot 256\cdot 3\cdot 8}{n_{eff} \cdot \log_2(973)}$ (in bits), and $n_{eff}$ is the effective sequence length as in Tab.~\ref{tab:diffusionhyper}.	%
  }
\end{figure*}
}

\newcommand{\modelsamplessmall}{
\begin{figure*}[thbp]
	\centering
    \renewcommand{\imwidth}{0.142\textwidth}
    \renewcommand{\impatha}[1]{img/cherries/cats/##1} 
    \renewcommand{\impathb}[1]{img/cherries/churches/##1} 
    \renewcommand{\impathc}[1]{img/cherries/beds/##1} 
    \renewcommand{\impathd}[1]{img/cherries/imagenet/##1} 
    \renewcommand{\impathe}[1]{img/cherries/ffhq/##1} 
    \setlength{\tabcolsep}{0pt}
    \renewcommand{\arraystretch}{0}
	\begin{small}
 	\begin{tabular}{ccccccc}
  	\toprule
	\includegraphics[width=\imwidth]{\impathe{00000002_0036}} &	        
	\includegraphics[width=\imwidth]{\impathe{00000000_0008}} &	         
	\includegraphics[width=\imwidth]{\impathe{00000004_0015}} &

	\includegraphics[width=\imwidth]{\impatha{0205_0069}} &	        
	\includegraphics[width=\imwidth]{\impatha{00000004_0010}} &	         
    \includegraphics[width=\imwidth]{\impatha{0204_0067}} &
    \includegraphics[width=\imwidth]{\impatha{0002}} \\

	\includegraphics[width=\imwidth]{\impathc{0001_0008}} &	         
	\includegraphics[width=\imwidth]{\impathc{0001_0007}} &	         
    \includegraphics[width=\imwidth]{\impathc{0024_0033}} &
    
    \includegraphics[width=\imwidth]{\impathb{0000_0010}} &	         
	\includegraphics[width=\imwidth]{\impathb{0002_0003}} &	        
	\includegraphics[width=\imwidth]{\impathb{0003_0012}} &	         
	\includegraphics[width=\imwidth]{\impathb{0004_0012}} \\

	\includegraphics[width=\imwidth]{\impathd{000004}} &	         
	\includegraphics[width=\imwidth]{\impathd{000001}} &	        
	\includegraphics[width=\imwidth]{\impathd{000002}} &	        
	\includegraphics[width=\imwidth]{\impathd{000008_1}} &	         
	\includegraphics[width=\imwidth]{\impathd{000012}} &	        
	\includegraphics[width=\imwidth]{\impathd{000026}} &	         
	\includegraphics[width=\imwidth]{\impathd{000034}} \\
  	\bottomrule
	\end{tabular}
	\end{small}
	\caption{\label{fig:modelsamplessmall} Samples from our models. Top row: FFHQ, LSUN-Cats, Middle row: LSUN-Bedrooms, LSUN-Churches, Bottom row: ImageNet.}
\end{figure*}
}

\newcommand{\ffhqinpainting}{
\begin{figure*}[thbp]
\centering
    \renewcommand{\imwidth}{0.09\textwidth}
 
    \renewcommand{\impatha}[1]{img/ffhqinpainting/##1} 
    \setlength{\tabcolsep}{0pt}		
    \renewcommand{\arraystretch}{0}
 	\begin{tabular}{ccccccccccc}
  	\toprule

	\includegraphics[align=c,width=\imwidth]{\impatha{masked_inputs_000_00001}} &	         
	\includegraphics[align=c,width=\imwidth]{\impatha{masked_inputs_000_00014}} &
	\includegraphics[align=c,width=\imwidth]{\impatha{masked_inputs_000_00000}} &	         
	\includegraphics[align=c,width=\imwidth]{\impatha{masked_inputs_000_00005}} &	         
	\includegraphics[align=c,width=\imwidth]{\impatha{masked_inputs_000_00006}} &	         
	\includegraphics[align=c,width=\imwidth]{\impatha{masked_inputs_000_00008}} &	         
	\includegraphics[align=c,width=\imwidth]{\impatha{masked_inputs_000_00009}} &
	\includegraphics[align=c,width=\imwidth]{\impatha{masked_inputs_000_00010}} &
	\includegraphics[align=c,width=\imwidth]{\impatha{masked_inputs_000_00011}} &
	\includegraphics[align=c,width=\imwidth]{\impatha{masked_inputs_000_00002}} &	         
	\includegraphics[align=c,width=\imwidth]{\impatha{masked_inputs_000_00018}}

  \\ 
	\includegraphics[align=c,width=\imwidth]{\impatha{samples_000_00001}} &	         
	\includegraphics[align=c,width=\imwidth]{\impatha{samples_005_00014}} &
	\includegraphics[align=c,width=\imwidth]{\impatha{samples_002_00000}} &	         
	\includegraphics[align=c,width=\imwidth]{\impatha{samples_000_00005}} &	         
	\includegraphics[align=c,width=\imwidth]{\impatha{samples_001_00006}} &	         
	\includegraphics[align=c,width=\imwidth]{\impatha{samples_002_00008}} &	         
	\includegraphics[align=c,width=\imwidth]{\impatha{samples_000_00009}} &
	\includegraphics[align=c,width=\imwidth]{\impatha{samples_002_00010}} &
	\includegraphics[align=c,width=\imwidth]{\impatha{samples_000_00011}} &
	\includegraphics[align=c,width=\imwidth]{\impatha{samples_003_00002}} &	         
	\includegraphics[align=c,width=\imwidth]{\impatha{samples_000_00018}}

  \\
	\includegraphics[align=c,width=\imwidth]{\impatha{samples_001_00001}} &	         
	\includegraphics[align=c,width=\imwidth]{\impatha{samples_006_00014}} &
	\includegraphics[align=c,width=\imwidth]{\impatha{samples_003_00000}} &	         
	\includegraphics[align=c,width=\imwidth]{\impatha{samples_006_00005}} &	         
	\includegraphics[align=c,width=\imwidth]{\impatha{samples_006_00006}} &	         
	\includegraphics[align=c,width=\imwidth]{\impatha{samples_007_00008}} &	         
	\includegraphics[align=c,width=\imwidth]{\impatha{samples_007_00009}} &
	\includegraphics[align=c,width=\imwidth]{\impatha{samples_007_00010}} &
	\includegraphics[align=c,width=\imwidth]{\impatha{samples_002_00011}} &
	\includegraphics[align=c,width=\imwidth]{\impatha{samples_005_00002}} &	         
	\includegraphics[align=c,width=\imwidth]{\impatha{samples_001_00018}}

  \\
  \bottomrule
	\end{tabular}
  
  \caption{\label{fig:localediting}Local editing application using markov chain of length 16 on FFHQ. By incorporating bidirectional context
  ImageBART is able to solve this unconditional inpainting task 
  (cf. Sec.~\ref{subsec:expthree}).}
\end{figure*}	
}

\newcommand{\conceptualcaptionsamples}{
\begin{figure}%
\centering
    \renewcommand{\imwidth}{0.11\textwidth}
    \renewcommand{\impatha}[1]{img/conceptuals/house/##1} 
    \renewcommand{\impathb}[1]{img/conceptuals/shapeofheart/##1} 
    \renewcommand{\impathc}[1]{img/conceptuals/mountains/##1} 
    \renewcommand{\impathd}[1]{img/conceptuals/texture/##1} 
	\setlength{\tabcolsep}{0pt}
    \renewcommand{\arraystretch}{0}
	\begin{small}
 	\begin{tabular}{ccc ccc ccc}
 	\multicolumn{3}{c}{\footnotesize{Mountains and hills \emph{\_\_\_}}} & \multicolumn{3}{c}{\footnotesize{\emph{\_\_\_} in the shape of a heart.}} & \multicolumn{3}{c}{\footnotesize{The texture of \emph{\_\_\_}}} \\  
\emph{\tiny{reflecting on a surface./}} & \emph{\tiny{covered in snow./}} & \emph{\tiny{during sunrise.}} & \emph{\tiny{An apple}} & \emph{\tiny{An avocado}}  & \emph{\tiny{Flames}} & \emph{\tiny{wood.}} & \emph{\tiny{pizza.}} & \emph{\tiny{water.}}\\
  	\toprule
  	\includegraphics[width=\imwidth]{\impathc{reflection/clip_0001}} &	         
	\includegraphics[width=\imwidth]{\impathc{snow/clip_0001}} &	        
	\includegraphics[width=\imwidth]{\impathc{sunrise/clip_0001}} &	         
	
	\includegraphics[width=\imwidth]{\impathb{apple/clip_0003}} &	         
	\includegraphics[width=\imwidth]{\impathb{avocado/subj_0001}} &	        
	\includegraphics[width=\imwidth]{\impathb{flames/clip_0002}} &
	
	\includegraphics[width=\imwidth]{\impathd{wood/clip_0001}} &	         
	\includegraphics[width=\imwidth]{\impathd{pizza/clip_0001}} &	        
	\includegraphics[width=\imwidth]{\impathd{water/water_0001}} \\

  	\includegraphics[width=\imwidth]{\impathc{reflection/clip_0006}} &	         
	\includegraphics[width=\imwidth]{\impathc{snow/clip_0002}} &	        
	\includegraphics[width=\imwidth]{\impathc{sunrise/clip_0002}} &	         
	
	\includegraphics[width=\imwidth]{\impathb{apple/clip_0002}} &	         
	\includegraphics[width=\imwidth]{\impathb{avocado/clip_0002}} &	        
	\includegraphics[width=\imwidth]{\impathb{flames/clip_0003}} &
	
	\includegraphics[width=\imwidth]{\impathd{wood/clip_0002}} &	         
	\includegraphics[width=\imwidth]{\impathd{pizza/clip_0002}} &	        
	\includegraphics[width=\imwidth]{\impathd{water/water_0002}} \\

  	\bottomrule
	\end{tabular}
	\end{small}
	\caption{\label{fig:conceptualcaptionsamples}Samples from text-conditional ImageBART. Best 2 of 32 with reranking as in \citep{DBLP:journals/corr/abs-2102-12092}. \vspace{-1em}}
\end{figure}	
}

\newcommand{\conditionalinpainting}{
\begin{figure*}[thbp]
\centering
    \renewcommand{\imwidth}{0.16\textwidth}
    \renewcommand{\impatha}[1]{img/editing/imagenet/##1} 
    \renewcommand{\impathb}[1]{img/editing/cc/##1} 
	\setlength{\tabcolsep}{0pt}
    \renewcommand{\arraystretch}{0}
	\begin{small}
 	\begin{tabular}{c@{\hskip 5pt}c@{\hskip 5pt}cccc}
 	Original & Masked & \multicolumn{4}{c}{Guidance} \\ 
  	\toprule
  	\tiny{\emph{Arctic fox (c279)}} & & \tiny{\emph{Lorikeet (c90)}} & \tiny{\emph{Zebra (c340)}} & \tiny{\emph{Tiger (c292)}} & \tiny{\emph{Green lizzard (c46)}} \\
  	\midrule
	\includegraphics[width=\imwidth]{\impatha{arctic_fox/original}} &	         
	\includegraphics[width=\imwidth]{\impatha{arctic_fox/mask}} &	 
	       
	\includegraphics[width=\imwidth]{\impatha{arctic_fox/arctic_lorikeet}} &	         
	\includegraphics[width=\imwidth]{\impatha{arctic_fox/zebra}} &   
	\includegraphics[width=\imwidth]{\impatha{arctic_fox/tiger}} &
	\includegraphics[width=\imwidth]{\impatha{arctic_fox/green_lizzard}} \\
	\midrule
	\tiny{\emph{'Man standing on a road'}} & & \tiny{\emph{'A beautiful sunset.'}} & \tiny{\emph{'Northern lights.'}} & \tiny{\emph{'The road leads straight}}  & \tiny{\emph{'A small city.'}} \\
	\tiny{\emph{'in nature in summer.'}}&&&&\tiny{\emph{to the coast.'}} & \\
  	\midrule	
	\includegraphics[width=\imwidth]{\impathb{original}} &	         
	\includegraphics[width=\imwidth]{\impathb{mask}} &	        

	\includegraphics[width=\imwidth]{\impathb{a_beautiful_sunset}} &	         
	\includegraphics[width=\imwidth]{\impathb{northern_lights}} &	
	\includegraphics[width=\imwidth]{\impathb{the_road_leads_straight_to_the_coast}} &	         
	\includegraphics[width=\imwidth]{\impathb{a_small_city}} \\
	
  	\bottomrule
	\end{tabular}
	\end{small}
  
	\caption{\label{fig:localcondediting}Conditionally guided inpainting results obtained from conditional ImageBART trained on the i) ImageNet (top row) and ii) Conceptual Captions (bottom row) datasets. \vspace{-1em}}
\end{figure*}	
}

\newcommand{\comparisonar}{
\begin{figure*}[thbp]
\centering
    \renewcommand{\imwidth}{0.11\textwidth}
 
    \renewcommand{\impatha}[1]{img/ar_comparison/bart/##1} 
    \renewcommand{\impathb}[1]{img/ar_comparison/tamingtk100/##1} 
    \renewcommand{\impathc}[1]{img/ar_comparison/bartchain/##1} 
    \renewcommand{\impathd}[1]{img/ar_comparison/tamingchaintk100/##1} 
    \setlength{\tabcolsep}{0pt}		
    \renewcommand{\arraystretch}{0}
 	\begin{tabular}{ccc@{\hskip 2pt}ccc@{\hskip 2pt}ccr@{}}
    \multicolumn{3}{c}{Masked Input} & \multicolumn{3}{c}{TT \citep{DBLP:journals/corr/abs-2012-09841}} & \multicolumn{3}{c}{ImageBART} \\ 
  	\midrule

	\includegraphics[align=c,width=\imwidth]{\impatha{masked_inputs_00094}} &	         
	\includegraphics[align=c,width=\imwidth]{\impatha{masked_inputs_00007}} &	         
	\includegraphics[align=c,width=\imwidth]{\impatha{masked_inputs_00091}} &	         
	
	\includegraphics[align=c,width=\imwidth]{\impathb{samples_00094}} &	         
	\includegraphics[align=c,width=\imwidth]{\impathb{samples_00007}} &	         
	\includegraphics[align=c,width=\imwidth]{\impathb{samples_00091}} &	         
	
	\includegraphics[align=c,width=\imwidth]{\impatha{samples_00094}} &	         
	\includegraphics[align=c,width=\imwidth]{\impatha{samples_00007}} &	         
	\includegraphics[align=c,width=\imwidth]{\impatha{samples_00091}}
	
	\\

  \midrule
    Input & \multicolumn{8}{c}{Iterative Refinement According to Global
    Context} \\

  \bottomrule
	\end{tabular}
    \renewcommand{\arraystretch}{1}
 	\begin{tabular}{c@{\hskip 2pt}cccccccc}

    \multirow{2}{*}{
      \parbox[c]{\imwidth}{\small \centering
      \vspace{-1em}
      ImageBART\vspace{0.25em}\\
      \includegraphics[align=c,width=\imwidth]{\impathc{masked_inputs_00023}}
      \vspace{0.25em}
      \\ TT \citep{DBLP:journals/corr/abs-2012-09841}
      }
    } &	         

	\includegraphics[align=c,width=\imwidth]{\impathc{sample_001_000_00023}} &	         
	\includegraphics[align=c,width=\imwidth]{\impathc{sample_000_000_00023}} &	         
	\includegraphics[align=c,width=\imwidth]{\impathc{sample_000_002_00023}} &	         
	\includegraphics[align=c,width=\imwidth]{\impathc{sample_000_004_00023}} &	         
	\includegraphics[align=c,width=\imwidth]{\impathc{sample_000_008_00023}} &	         
	\includegraphics[align=c,width=\imwidth]{\impathc{sample_000_010_00023}} &	         
	\includegraphics[align=c,width=\imwidth]{\impathc{sample_000_012_00023}} &	         
	\includegraphics[align=c,width=\imwidth]{\impathc{sample_000_015_00023}}
	\\

    &

	\includegraphics[align=c,width=\imwidth]{\impathd{sample_000_000_00023}} &	         
	\includegraphics[align=c,width=\imwidth]{\impathd{sample_000_001_00023}} &	         
	\includegraphics[align=c,width=\imwidth]{\impathd{sample_000_002_00023}} &	         
	\includegraphics[align=c,width=\imwidth]{\impathd{sample_000_004_00023}} &	         
	\includegraphics[align=c,width=\imwidth]{\impathd{sample_000_008_00023}} &	         
	\includegraphics[align=c,width=\imwidth]{\impathd{sample_000_010_00023}} &	         
	\includegraphics[align=c,width=\imwidth]{\impathd{sample_000_012_00023}} &	         
	\includegraphics[align=c,width=\imwidth]{\impathd{sample_000_015_00023}}
	\\
  \midrule
	\end{tabular}

	\caption{\label{fig:arcomp} Without global context, AR models fail at completing upper
  halfs, contrasting ImageBART.\vspace{-1em}}
  
\end{figure*}	
}

\newcommand{\suppcomparisonar}{
\begin{figure*}[thbp]
\centering
    \renewcommand{\imwidth}{0.11\textwidth}
 
    \renewcommand{\impatha}[1]{img/ar_comparison/bart/##1} 
    \renewcommand{\impathb}[1]{img/ar_comparison/tamingtk100/##1} 
    \renewcommand{\impathc}[1]{img/ar_comparison/bartchain/##1} 
    \renewcommand{\impathd}[1]{img/ar_comparison/tamingchaintk100/##1} 
    \setlength{\tabcolsep}{0pt}		
    \renewcommand{\arraystretch}{0}
 	\begin{tabular}{ccc@{\hskip 2pt}ccc@{\hskip 2pt}ccr@{}}
    \multicolumn{3}{c}{Masked Input} & \multicolumn{3}{c}{TT \citep{DBLP:journals/corr/abs-2012-09841}} & \multicolumn{3}{c}{ImageBART} \\ 
  	\midrule

	\includegraphics[align=c,width=\imwidth]{\impatha{masked_inputs_00087}} &	         
	\includegraphics[align=c,width=\imwidth]{\impatha{masked_inputs_00093}} &	         
	\includegraphics[align=c,width=\imwidth]{\impatha{masked_inputs_00086}} &	         
	
	\includegraphics[align=c,width=\imwidth]{\impathb{samples_00087}} &	         
	\includegraphics[align=c,width=\imwidth]{\impathb{samples_00093}} &	         
	\includegraphics[align=c,width=\imwidth]{\impathb{samples_00086}} &	         
	
	\includegraphics[align=c,width=\imwidth]{\impatha{samples_00087}} &	         
	\includegraphics[align=c,width=\imwidth]{\impatha{samples_00093}} &	         
	\includegraphics[align=c,width=\imwidth]{\impatha{samples_00086}}
	
	\\

	\includegraphics[align=c,width=\imwidth]{\impatha{masked_inputs_00080}} &	         
	\includegraphics[align=c,width=\imwidth]{\impatha{masked_inputs_00071}} &	         
	\includegraphics[align=c,width=\imwidth]{\impatha{masked_inputs_00027}} &	         
	
  \includegraphics[align=c,width=\imwidth]{\impathb{samples_00080}} &	         
	\includegraphics[align=c,width=\imwidth]{\impathb{samples_00071}} &	         
	\includegraphics[align=c,width=\imwidth]{\impathb{samples_00027}} &	         
	
	\includegraphics[align=c,width=\imwidth]{\impatha{samples_00080}} &	         
	\includegraphics[align=c,width=\imwidth]{\impatha{samples_00071}} &	         
	\includegraphics[align=c,width=\imwidth]{\impatha{samples_00027}}

  \\ 

	\includegraphics[align=c,width=\imwidth]{\impatha{masked_inputs_00079}} &	         
	\includegraphics[align=c,width=\imwidth]{\impatha{masked_inputs_00068}} &	         
	\includegraphics[align=c,width=\imwidth]{\impatha{masked_inputs_00067}} &	         
	
  \includegraphics[align=c,width=\imwidth]{\impathb{samples_00079}} &	         
	\includegraphics[align=c,width=\imwidth]{\impathb{samples_00068}} &	         
	\includegraphics[align=c,width=\imwidth]{\impathb{samples_00067}} &	         
	
	\includegraphics[align=c,width=\imwidth]{\impatha{samples_00079}} &	         
	\includegraphics[align=c,width=\imwidth]{\impatha{samples_00068}} &	         
	\includegraphics[align=c,width=\imwidth]{\impatha{samples_00067}}

  \\ 
	
	\includegraphics[align=c,width=\imwidth]{\impatha{masked_inputs_00066}} &	         
	\includegraphics[align=c,width=\imwidth]{\impatha{masked_inputs_00065}} &	         
	\includegraphics[align=c,width=\imwidth]{\impatha{masked_inputs_00051}} &	         
	
  \includegraphics[align=c,width=\imwidth]{\impathb{samples_00066}} &	         
	\includegraphics[align=c,width=\imwidth]{\impathb{samples_00065}} &	         
	\includegraphics[align=c,width=\imwidth]{\impathb{samples_00051}} &	         
	
	\includegraphics[align=c,width=\imwidth]{\impatha{samples_00066}} &	         
	\includegraphics[align=c,width=\imwidth]{\impatha{samples_00065}} &	         
	\includegraphics[align=c,width=\imwidth]{\impatha{samples_00051}}

  \\ 

	\includegraphics[align=c,width=\imwidth]{\impatha{masked_inputs_00049}} &	         
	\includegraphics[align=c,width=\imwidth]{\impatha{masked_inputs_00035}} &	         
	\includegraphics[align=c,width=\imwidth]{\impatha{masked_inputs_00040}} &	         
	
  \includegraphics[align=c,width=\imwidth]{\impathb{samples_00049}} &	         
	\includegraphics[align=c,width=\imwidth]{\impathb{samples_00035}} &	         
	\includegraphics[align=c,width=\imwidth]{\impathb{samples_00040}} &	         
	
	\includegraphics[align=c,width=\imwidth]{\impatha{samples_00049}} &	         
	\includegraphics[align=c,width=\imwidth]{\impatha{samples_00035}} &	         
	\includegraphics[align=c,width=\imwidth]{\impatha{samples_00040}}

  \\ 

	\includegraphics[align=c,width=\imwidth]{\impatha{masked_inputs_00009}} &	         
	\includegraphics[align=c,width=\imwidth]{\impatha{masked_inputs_00008}} &	         
	\includegraphics[align=c,width=\imwidth]{\impatha{masked_inputs_00001}} &	         
	
	\includegraphics[align=c,width=\imwidth]{\impathb{samples_00009}} &	         
	\includegraphics[align=c,width=\imwidth]{\impathb{samples_00008}} &	         
	\includegraphics[align=c,width=\imwidth]{\impathb{samples_00001}} &	         
	
	\includegraphics[align=c,width=\imwidth]{\impatha{samples_00009}} &	         
	\includegraphics[align=c,width=\imwidth]{\impatha{samples_00008}} &	         
	\includegraphics[align=c,width=\imwidth]{\impatha{samples_00001}} \\
  \midrule
    Input & \multicolumn{8}{c}{$\xrightarrow{\text{\hspace{5em} Iterative Refinement According to Global
    Context \hspace{5em}}}$} \\

  \bottomrule
	\end{tabular}
    \renewcommand{\arraystretch}{1}
 	\begin{tabular}{c@{\hskip 2pt}cccccccc}

    \multirow{2}{*}{
      \parbox[c]{\imwidth}{\small \centering
      \vspace{-1em}
      ImageBART\vspace{0.25em}\\
      \includegraphics[align=c,width=\imwidth]{\impathc{masked_inputs_00006}}
      \vspace{0.25em}
      \\ TT \citep{DBLP:journals/corr/abs-2012-09841}
      }
    } &	         

	\includegraphics[align=c,width=\imwidth]{\impathc{sample_001_000_00006}} &	         
	\includegraphics[align=c,width=\imwidth]{\impathc{sample_000_000_00006}} &	         
	\includegraphics[align=c,width=\imwidth]{\impathc{sample_000_002_00006}} &	         
	\includegraphics[align=c,width=\imwidth]{\impathc{sample_000_004_00006}} &	         
	\includegraphics[align=c,width=\imwidth]{\impathc{sample_000_008_00006}} &	         
	\includegraphics[align=c,width=\imwidth]{\impathc{sample_000_010_00006}} &	         
	\includegraphics[align=c,width=\imwidth]{\impathc{sample_000_012_00006}} &	         
	\includegraphics[align=c,width=\imwidth]{\impathc{sample_000_015_00006}}
	\\

    &

	\includegraphics[align=c,width=\imwidth]{\impathd{sample_000_000_00006}} &	         
	\includegraphics[align=c,width=\imwidth]{\impathd{sample_000_001_00006}} &	         
	\includegraphics[align=c,width=\imwidth]{\impathd{sample_000_002_00006}} &	         
	\includegraphics[align=c,width=\imwidth]{\impathd{sample_000_004_00006}} &	         
	\includegraphics[align=c,width=\imwidth]{\impathd{sample_000_008_00006}} &	         
	\includegraphics[align=c,width=\imwidth]{\impathd{sample_000_010_00006}} &	         
	\includegraphics[align=c,width=\imwidth]{\impathd{sample_000_012_00006}} &	         
	\includegraphics[align=c,width=\imwidth]{\impathd{sample_000_015_00006}}
	\\
  \midrule

    \multirow{2}{*}{
      \parbox[c]{\imwidth}{\small \centering
      \vspace{-1em}
      ImageBART\vspace{0.25em}\\
      \includegraphics[align=c,width=\imwidth]{\impathc{masked_inputs_00002}}
      \vspace{0.25em}
      \\ TT \citep{DBLP:journals/corr/abs-2012-09841}
      }
    } &	         

	\includegraphics[align=c,width=\imwidth]{\impathc{sample_001_000_00002}} &	         
	\includegraphics[align=c,width=\imwidth]{\impathc{sample_000_000_00002}} &	         
	\includegraphics[align=c,width=\imwidth]{\impathc{sample_000_002_00002}} &	         
	\includegraphics[align=c,width=\imwidth]{\impathc{sample_000_004_00002}} &	         
	\includegraphics[align=c,width=\imwidth]{\impathc{sample_000_008_00002}} &	         
	\includegraphics[align=c,width=\imwidth]{\impathc{sample_000_010_00002}} &	         
	\includegraphics[align=c,width=\imwidth]{\impathc{sample_000_012_00002}} &	         
	\includegraphics[align=c,width=\imwidth]{\impathc{sample_000_015_00002}}
	\\

    &

	\includegraphics[align=c,width=\imwidth]{\impathd{sample_000_000_00002}} &	         
	\includegraphics[align=c,width=\imwidth]{\impathd{sample_000_001_00002}} &	         
	\includegraphics[align=c,width=\imwidth]{\impathd{sample_000_002_00002}} &	         
	\includegraphics[align=c,width=\imwidth]{\impathd{sample_000_004_00002}} &	         
	\includegraphics[align=c,width=\imwidth]{\impathd{sample_000_008_00002}} &	         
	\includegraphics[align=c,width=\imwidth]{\impathd{sample_000_010_00002}} &	         
	\includegraphics[align=c,width=\imwidth]{\impathd{sample_000_012_00002}} &	         
	\includegraphics[align=c,width=\imwidth]{\impathd{sample_000_015_00002}}
	\\
  \midrule
    \multirow{2}{*}{
      \parbox[c]{\imwidth}{\small \centering
      \vspace{-1em}
      ImageBART\vspace{0.25em}\\
      \includegraphics[align=c,width=\imwidth]{\impathc{masked_inputs_00003}}
      \vspace{0.25em}
      \\ TT \citep{DBLP:journals/corr/abs-2012-09841}
      }
    } &
	\includegraphics[align=c,width=\imwidth]{\impathc{sample_001_000_00003}} &	         
	\includegraphics[align=c,width=\imwidth]{\impathc{sample_000_000_00003}} &	         
	\includegraphics[align=c,width=\imwidth]{\impathc{sample_000_002_00003}} &	         
	\includegraphics[align=c,width=\imwidth]{\impathc{sample_000_004_00003}} &	         
	\includegraphics[align=c,width=\imwidth]{\impathc{sample_000_008_00003}} &	         
	\includegraphics[align=c,width=\imwidth]{\impathc{sample_000_010_00003}} &	         
	\includegraphics[align=c,width=\imwidth]{\impathc{sample_000_012_00003}} &	         
	\includegraphics[align=c,width=\imwidth]{\impathc{sample_000_015_00003}}
  \\
    &
	\includegraphics[align=c,width=\imwidth]{\impathd{sample_000_000_00003}} &	         
	\includegraphics[align=c,width=\imwidth]{\impathd{sample_000_001_00003}} &	         
	\includegraphics[align=c,width=\imwidth]{\impathd{sample_000_002_00003}} &	         
	\includegraphics[align=c,width=\imwidth]{\impathd{sample_000_007_00003}} &	         
	\includegraphics[align=c,width=\imwidth]{\impathd{sample_000_008_00003}} &	         
	\includegraphics[align=c,width=\imwidth]{\impathd{sample_000_010_00003}} &	         
	\includegraphics[align=c,width=\imwidth]{\impathd{sample_000_013_00003}} &	         
	\includegraphics[align=c,width=\imwidth]{\impathd{sample_000_003_00003}}
  \\
	\end{tabular}

	\caption{\label{fig:supparcomp} Additional examples for upper half completion
  as in Fig.~\ref{fig:arcomp}. The top shows masked inputs, results by TT \citep{DBLP:journals/corr/abs-2012-09841}
  and results by ImageBART. The bottom shows every other sample of the
  forward-backward chain described in Sec.~\ref{subsec:expthree} and Sec.~\ref{supp:maskeddenoising}.
  ImageBART can incorporate global context to produce consistent
  completions, whereas TT is limited to context from above and thus fails to
  produce consistent completions.}
  
\end{figure*}	
}

%% file: tables.tex
\newcommand{\tablation}{
\begin{table}%
\centering
\begin{footnotesize}
  \begin{adjustbox}{max width=\linewidth}
  \begin{tabular}{lcc}
  \toprule
  \multicolumn{3}{c}{Unconditional Generation} \\
  \midrule
method  & FID $\downarrow$ & IS $\uparrow$ \\
\midrule
 TT ($T=2$) & 12.44 & $3.98 \pm 0.07$ \\
 ImageBART ($T=3$) & 12.55 & $3.98 \pm 0.07$ \\
 ImageBART ($T=5$) & 10.69 & $4.27 \pm 0.05$ \\
 ImageBART ($T=9$) & 10.81 & $4.49 \pm 0.05$ \\
 \bottomrule
  \end{tabular}
    \begin{tabular}{lcc}
  \toprule
    \multicolumn{3}{c}{Upper Half Completion} \\
    \midrule
method  & FID $\downarrow$ & IS $\uparrow$ \\
\midrule
 TT ($T=2$) & 11.80 & $4.48 \pm 0.10$ \\
 ImageBART ($T=3$) & 9.25 & $4.49 \pm 0.13$ \\
 ImageBART ($T=5$) & 6.87 & $4.81 \pm 0.13$ \\
 ImageBART ($T=9$) & 6.64 & $4.86 \pm 0.15$ \\
  \bottomrule
  \end{tabular}
  
  \end{adjustbox}
\end{footnotesize}
  \caption{\label{tab:tablation}
Assessing the effect of different $T$ with a fixed number of parameters
  distributed equally over all scales. All models are trained on FFHQ. \emph{Left:} Full image generation results. \emph{Right:} Using the example
  of upper image completion, we evaluate the ability to complete and modifiy an
  image, see Sec.~\ref{subsec:expthree} and \ref{subsec:expfour}.
  }
\end{table}
}

\newcommand{\fids}{
\begin{table}[t]
\centering
\begin{footnotesize}
\centering
    \setlength{\tabcolsep}{1pt}%
  \resizebox{.45\textwidth}{!}{
  \begin{tabular}{lccccccc}
    \toprule
    Method & Cats & & Beds & & Churches & & FFHQ \\
    \midrule
    VDVAE~\cite{DBLP:journals/corr/abs-2011-10650} & -- & & -- & & -- & & 28.5\\
    \midrule
    DDPM~\cite{DBLP:conf/nips/HoJA20} & 19.75 & & 4.90 & & 7.89 &&--\\
    \midrule
  	StyleGAN2~\cite{DBLP:journals/corr/abs-1912-04958} & 7.25 & & 2.35 & & 3.86 &&3.8\\
	BigGAN~\cite{big_gan_brock} & -- & & -- & & -- &&12.4\\
	\midrule
	DCT \cite{DBLP:journals/corr/abs-2103-03841} & -- & & 6.40 & & 7.56 && 13.06 \\    
    TT~\citep{DBLP:journals/corr/abs-2012-09841} & 17.31 & & 6.35 & & 7.81 &&11.4\\
	\midrule
    ImageBART  & 15.09 & &  5.51 & & 7.32 & & 9.57 	\\
  \bottomrule
  \end{tabular}%
  }
    \hfill%
  \parbox[b]{0.54\textwidth}{
\centering
\small
    \renewcommand{\imwidth}{0.047\textwidth}
    \renewcommand{\impatha}[1]{img/diffusion_comparison/ssde/##1} 
	\renewcommand{\impathb}[1]{img/diffusion_comparison/ours/##1} 
    \renewcommand{\impathc}[1]{img/diffusion_comparison/ddpm/##1} 
	\setlength{\tabcolsep}{0pt}		
 	\begin{tabular}{c@{\hskip 3pt}ccc@{\hskip 2pt}ccc@{\hskip 2pt}ccr@{}}
    \toprule
 	 & \multicolumn{3}{c}{ImageBART} & \multicolumn{3}{c}{DDPM} & \multicolumn{3}{c}{SSDE} \\ 
  	\toprule
	\emph{Churches} &		
	\includegraphics[align=c,width=\imwidth]{\impathb{churches/0001_0015}} &	         
	\includegraphics[align=c,width=\imwidth]{\impathb{churches/00000001_0171}} &	        
	\includegraphics[align=c,width=\imwidth]{\impathb{churches/0002_0015}} &	         
	
	\includegraphics[align=c,width=\imwidth]{\impathc{church/sample2}} &	         
	\includegraphics[align=c,width=\imwidth]{\impathc{church/sample3}} &	        
	\includegraphics[align=c,width=\imwidth]{\impathc{church/sample1}} &
	
	\includegraphics[align=c,width=\imwidth]{\impatha{church/sample-8}} &	         
	\includegraphics[align=c,width=\imwidth]{\impatha{church/sample-513}} &	        
	\includegraphics[align=c,width=\imwidth]{\impatha{church/sample-514}} 
	\\

	\emph{Cats} &
	\includegraphics[align=c,width=\imwidth]{\impathb{cats/00000001_0014}} &	         
	\includegraphics[align=c,width=\imwidth]{\impathb{cats/00000007_0020}} &	        
	\includegraphics[align=c,width=\imwidth]{\impathb{cats/00000005_0020}} & 	
	
	\includegraphics[align=c,width=\imwidth]{\impathc{cats/Selection_122}} &	         
	\includegraphics[align=c,width=\imwidth]{\impathc{cats/Selection_123}} &	        
	\includegraphics[align=c,width=\imwidth]{\impathc{cats/Selection_124}} &
	
	\includegraphics[align=c,width=\imwidth]{\impatha{cats/sample-524}} &	         
	\includegraphics[align=c,width=\imwidth]{\impatha{cats/sample-1024}} &	        
	\includegraphics[align=c,width=\imwidth]{\impatha{cats/sample-1026}}  \\

	\emph{cIN (c14)} &
	\includegraphics[align=c,width=\imwidth]{\impathb{imagenet/indigo_bird_14/000001}} &	         
	\includegraphics[align=c,width=\imwidth]{\impathb{imagenet/indigo_bird_14/000002}} &	        
	\includegraphics[align=c,width=\imwidth]{\impathb{imagenet/indigo_bird_14/000006}} &
	
	\includegraphics[align=c,width=\imwidth]{\impathc{imagenet/indigo_bird_14/sample-15}} &	         
	\includegraphics[align=c,width=\imwidth]{\impathc{imagenet/indigo_bird_14/sample-81}} &	        
	\includegraphics[align=c,width=\imwidth]{\impathc{imagenet/indigo_bird_14/sample-114}} & 
	
	\includegraphics[align=c,width=\imwidth]{\impatha{imagenet/indigo_bird_14/sample-6}} &	         
	\includegraphics[align=c,width=\imwidth]{\impatha{imagenet/indigo_bird_14/sample-25}} &	        
	\includegraphics[align=c,width=\imwidth]{\impatha{imagenet/indigo_bird_14/sample-70}}  \\

	\emph{cIN (c323)} &
	\includegraphics[align=c,width=\imwidth]{\impathb{imagenet/monarch_butterfly_323/000001}} &	         
	\includegraphics[align=c,width=\imwidth]{\impathb{imagenet/monarch_butterfly_323/000005}} &	        
	\includegraphics[align=c,width=\imwidth]{\impathb{imagenet/monarch_butterfly_323/000007}} &
	
	\includegraphics[align=c,width=\imwidth]{\impathc{imagenet/monarch_butterfly_323/sample-142}} &	         
	\includegraphics[align=c,width=\imwidth]{\impathc{imagenet/monarch_butterfly_323/sample-208}} &	        
	\includegraphics[align=c,width=\imwidth]{\impathc{imagenet/monarch_butterfly_323/sample-220}} &
	
	\includegraphics[align=c,width=\imwidth]{\impatha{imagenet/monarch_butterfly_323/sample-63}} &	         
	\includegraphics[align=c,width=\imwidth]{\impatha{imagenet/monarch_butterfly_323/sample-71}} &	        
	\includegraphics[align=c,width=\imwidth]{\impatha{imagenet/monarch_butterfly_323/sample-96}}  \\		
	
	\emph{cIN (c963)} &
	\includegraphics[align=c,width=\imwidth]{\impathb{imagenet/pizza_963/000002}} &	         
	\includegraphics[align=c,width=\imwidth]{\impathb{imagenet/pizza_963/000004}} &	        
	\includegraphics[align=c,width=\imwidth]{\impathb{imagenet/pizza_963/000007}} &
	
	\includegraphics[align=c,width=\imwidth]{\impathc{imagenet/pizza_963/sample-163}} &	         
	\includegraphics[align=c,width=\imwidth]{\impathc{imagenet/pizza_963/sample-262}} &	        
	\includegraphics[align=c,width=\imwidth]{\impathc{imagenet/pizza_963/sample-405}} &
	
	\includegraphics[align=c,width=\imwidth]{\impatha{imagenet/pizza_963/sample-3}} &	         
	\includegraphics[align=c,width=\imwidth]{\impatha{imagenet/pizza_963/sample-9}} &	        
	\includegraphics[align=c,width=\imwidth]{\impatha{imagenet/pizza_963/sample-27}}  \\	
  	\bottomrule
	\end{tabular}
    }
\end{footnotesize}
  \caption{\label{tab:fids} \emph{Left:} FIDs on the
  LSUN-\{Churches,Beds,Cats\}~\citep{DBLP:journals/corr/YuZSSX15} and
  FFHQ~\cite{stylegan} datasets.
  \emph{Right:} Corresponding qualitative comparisons. Qualitative comparisons with TT can be found in 
Fig.~\ref{fig:cincomp} and Fig.~\ref{fig:comptext2imgsupp}
  \vspace{-1.5em}}
\end{table}
}

\newcommand{\quantcond}{
\begin{table}[t]
\centering
\begin{footnotesize}
\parbox[b]{0.49\textwidth}{%
\centering
    \setlength{\tabcolsep}{4pt}%
  \begin{tabular}[t]{lcccc}
    \toprule
 	 & \multicolumn{4}{c}{rejection rate for cIN sampling} \\ 
    \cmidrule{2-5}
      & 1.0 & 0.5 & 0.25 & 0.05 \\
    \midrule
    FID & 21.19 & 13.12 & 9.77 & 7.44 \\
    IS & 61.6\tiny$\pm 0.8$ & 109.5\tiny$\pm 2.3$ & 146.2\tiny$\pm 3.8$ & 273.5\tiny$\pm 4.1$ \\
  \bottomrule
\end{tabular}
}%
\hspace{2mm}
\parbox[b]{.49\textwidth}{%
    \setlength{\tabcolsep}{.5em}%
  \begin{tabular}[t]{lccc}
    \toprule
 	 \multicolumn{4}{c}{Text-conditional image synthesis on CC~\citep{sharma2018conceptual}}\\ 
    \midrule
    Method  & FID $\downarrow$ & IS $\downarrow$ & CLIP-score $\uparrow$\\
    \midrule
    TT~\citep{DBLP:journals/corr/abs-2012-09841} & 28.86 & 13.11\tiny$\pm 0.43$ & 0.20\tiny$\pm 0.03$\\
    ImageBART & 22.61 & 15.27\tiny$\pm 0.59$ & 0.23\tiny$\pm 0.03$ \\
  \bottomrule
  \hfill
\end{tabular}
}%
\end{footnotesize}
\caption{\label{fig:quant_cond1} Quantitative analysis on conditional models.
  Left: Results on class conditional Imagenet for different rejection rates, see also Fig,~\ref{fig:cincomp} in the supplemental.
  Right: Results of text-conditional ImageBART and comparison with TT~\citep{DBLP:journals/corr/abs-2012-09841} on the CC test set. 
  Corresponding qualitative comparisons can be found in Fig.~\ref{fig:comptext2imgsupp}.
  \vspace{-1.5em}
  }
\end{table}
}

%% file: figures_supplementary.tex
\providecommand{\imwidth}{}
\providecommand{\imheight}{}
\providecommand{\colwidth}{}
\providecommand{\cmidrulewidth}{}

\providecommand{\impaths}[1]{}

\newcommand{\churchessamplessupp}{
\begin{figure*}[thbp]
	\centering
    \renewcommand{\imwidth}{\textwidth}
    \renewcommand{\impaths}[1]{img/cherries/churches/supp/##1} 
    \setlength{\tabcolsep}{0pt}
    \renewcommand{\arraystretch}{0}
	\begin{small}
 	\begin{tabular}{cccccc}
  	\toprule
	         
	\includegraphics[width=\imwidth]{\impaths{summary_grid}} \\

  	\bottomrule
	\end{tabular}
	\end{small}
	\caption{\label{fig:churchessamplessupp} Additional $256 \times 256$ samples on the LSUN-church dataset.}
\end{figure*}
}

\newcommand{\bedroomsgridsupp}{
\begin{figure*}[thbp]
	\centering
    \renewcommand{\imwidth}{\textwidth}
    \renewcommand{\impaths}[1]{img/cherries/beds/##1} 
	\includegraphics[width=\imwidth]{\impaths{grid1.jpg}}
	\caption{\label{fig:bedroomsgridsupp} Additional random samples from our model trained on the LSUN-Bedrooms dataset.}
\end{figure*}
}

\newcommand{\catsgridsupp}{
\begin{figure*}[thbp]
	\centering
    \renewcommand{\imwidth}{\textwidth}
    \renewcommand{\impaths}[1]{img/cherries/cats/##1} 
	\includegraphics[width=\imwidth]{\impaths{grid1.jpg}}
	\caption{\label{fig:catsgridsupp} Additional random samples from our model trained on the LSUN-Cats dataset.}
\end{figure*}
}

\newcommand{\ffhqsamplessupp}{
\begin{figure*}[thbp]
	\centering
    \renewcommand{\imwidth}{\textwidth}
    \renewcommand{\impaths}[1]{img/cherries/ffhq/supp/##1} 
    \setlength{\tabcolsep}{0pt}
    \renewcommand{\arraystretch}{0}
	\begin{small}
 	\begin{tabular}{cccccc}
  	\toprule
	\includegraphics[width=\imwidth]{\impaths{summary_grid.jpg}} \\
  	\bottomrule
	\end{tabular}
	\end{small}
	\caption{\label{fig:ffhqsamplessupp} Additional $256 \times 256$ samples on the FFHQ dataset}
\end{figure*}
}

\newcommand{\sflckrconditionallarge}{
\begin{figure*}[thbp]
	\centering
    \renewcommand{\imwidth}{0.48\textwidth}
    \renewcommand{\imheight}{0.2\textheight}
    \renewcommand{\impaths}[1]{img/conditional/sflckr/##1} 
    \setlength{\tabcolsep}{0pt}
    \renewcommand{\arraystretch}{0}
	\begin{small}
 	\begin{tabular}{cc}
  	\toprule         
	\includegraphics[height=.2025\textheight]{\impaths{sem_syn_map1}} &	        
	\includegraphics[height=.2025\textheight]{\impaths{sem_syn_map2}} \\
	         
	\includegraphics[height=\imheight]{\impaths{sem_syn_exmpl1_1}} &	        
	\includegraphics[height=\imheight]{\impaths{sem_syn_exmpl2_1}} \\
         
	\includegraphics[height=\imheight]{\impaths{sem_syn_exmpl1_2}} &	        
	\includegraphics[height=\imheight]{\impaths{sem_syn_exmpl2_2}} \\
	
	\includegraphics[height=\imheight]{\impaths{sem_syn_exmpl1_3}} &	        
	\includegraphics[height=\imheight]{\impaths{sem_syn_exmpl2_3}} \\

  	\bottomrule
	\end{tabular}
	\end{small}
	\caption{\label{fig:sflickrsamplesone} Additional samples on semantic image synthesis. Left: $1024 \times 656$ pix. Right: $1024 \times 608$ pix.}
\end{figure*}
}

\newcommand{\imagenetsamplessupp}{
\begin{figure*}[thbp]
	\centering
    \renewcommand{\imwidth}{0.16\textwidth}
    \renewcommand{\impaths}[1]{img/cherries/imagenet/##1} 
    \setlength{\tabcolsep}{0pt}
    \renewcommand{\arraystretch}{0}
	\begin{small}
 	\begin{tabular}{cccccc}
  	\toprule
	\includegraphics[width=\imwidth]{\impaths{000001}} &	         
    \includegraphics[width=\imwidth]{\impaths{000002_2}} &	         
	\includegraphics[width=\imwidth]{\impaths{000003}} &	        
	\includegraphics[width=\imwidth]{\impaths{000003_1}} &	         
	\includegraphics[width=\imwidth]{\impaths{000003_2}} &	        
	\includegraphics[width=\imwidth]{\impaths{000003_3}} \\
	
	\includegraphics[width=\imwidth]{\impaths{000001_3}} &	         
    \includegraphics[width=\imwidth]{\impaths{000003_5}} &	         
	\includegraphics[width=\imwidth]{\impaths{000004}} &	        
	\includegraphics[width=\imwidth]{\impaths{000004_1}} &	         
	\includegraphics[width=\imwidth]{\impaths{000004_2}} &	        
	\includegraphics[width=\imwidth]{\impaths{000004_3}} \\

	\includegraphics[width=\imwidth]{\impaths{000005}} &	         
    \includegraphics[width=\imwidth]{\impaths{000005_1}} &	         
	\includegraphics[width=\imwidth]{\impaths{000007}} &	        
	\includegraphics[width=\imwidth]{\impaths{000008}} &	         
	\includegraphics[width=\imwidth]{\impaths{000006}} &	        
	\includegraphics[width=\imwidth]{\impaths{000007_2}} \\
	
	\includegraphics[width=\imwidth]{\impaths{000001_1}} &	         
    \includegraphics[width=\imwidth]{\impaths{000016}} &	         
	\includegraphics[width=\imwidth]{\impaths{000026}} &	        
	\includegraphics[width=\imwidth]{\impaths{000034}} &	         
	\includegraphics[width=\imwidth]{\impaths{000043}} &	        
	\includegraphics[width=\imwidth]{\impaths{000001_2}} \\

	\includegraphics[width=\imwidth]{\impaths{000003_4}} &	         
    \includegraphics[width=\imwidth]{\impaths{000001_4}} &	         
	\includegraphics[width=\imwidth]{\impaths{000038}} &	        
	\includegraphics[width=\imwidth]{\impaths{000002_4}} &	         
	\includegraphics[width=\imwidth]{\impaths{000004_4}} &	        
	\includegraphics[width=\imwidth]{\impaths{000005_2}} \\

	\includegraphics[width=\imwidth]{\impaths{000005_3}} &	         
    \includegraphics[width=\imwidth]{\impaths{000011}} &	         
	\includegraphics[width=\imwidth]{\impaths{000012}} &	        
	\includegraphics[width=\imwidth]{\impaths{000017}} &	         
	\includegraphics[width=\imwidth]{\impaths{000034_1}} &	        
	\includegraphics[width=\imwidth]{\impaths{000002_3}} \\
  	\bottomrule
	\end{tabular}
	\end{small}
	\caption{\label{fig:imagenetsamplessupp} Additional samples for class-conditional synthesis results on ImageNet.
  }
\end{figure*}
}

\newcommand{\texttoimgcomptt}{
\begin{figure*}[thbp]
\centering
\renewcommand{\colwidth}{0.113\textwidth}
\renewcommand{\cmidrulewidth}{1pt}
\setlength{\tabcolsep}{1pt}
\begin{footnotesize}
\begin{tabular}{m{\colwidth}  m{\colwidth}  m{\colwidth}  m{\colwidth}  m{\colwidth} m{\colwidth} | m{\colwidth} m{\colwidth}}
\cmidrule[\cmidrulewidth]{1-8}
\tiny\emph{~ Sunset over the}
& \tiny\emph{~ Map of the world}
& \tiny\emph{~~ Crowded scene}
& \tiny\emph{~~~ A small house}
& \multicolumn{2}{c}{\tiny\emph{A photograph of a}}
& \multicolumn{2}{c}{\tiny\emph{A vector illustration of}}\\
\tiny\emph{~ skyline of a city.}
& \tiny\emph{~ in the year 2077.}
& \tiny\emph{~~in front of a pub.}
& \tiny\emph{~~in the wilderness.}
& \quad ~ \tiny\emph{beach.}
& \tiny\emph{crowd of people.}
& \quad ~~~ \tiny\emph{a tree.}
& \quad ~ \tiny\emph{the brain.}  \\ 
\cmidrule[\cmidrulewidth]{1-8}
\multicolumn{8}{c}{\small{ImageBart}} \\
\cmidrule[\cmidrulewidth]{1-8}
\multicolumn{8}{c}{\includegraphics[width=.95\textwidth]{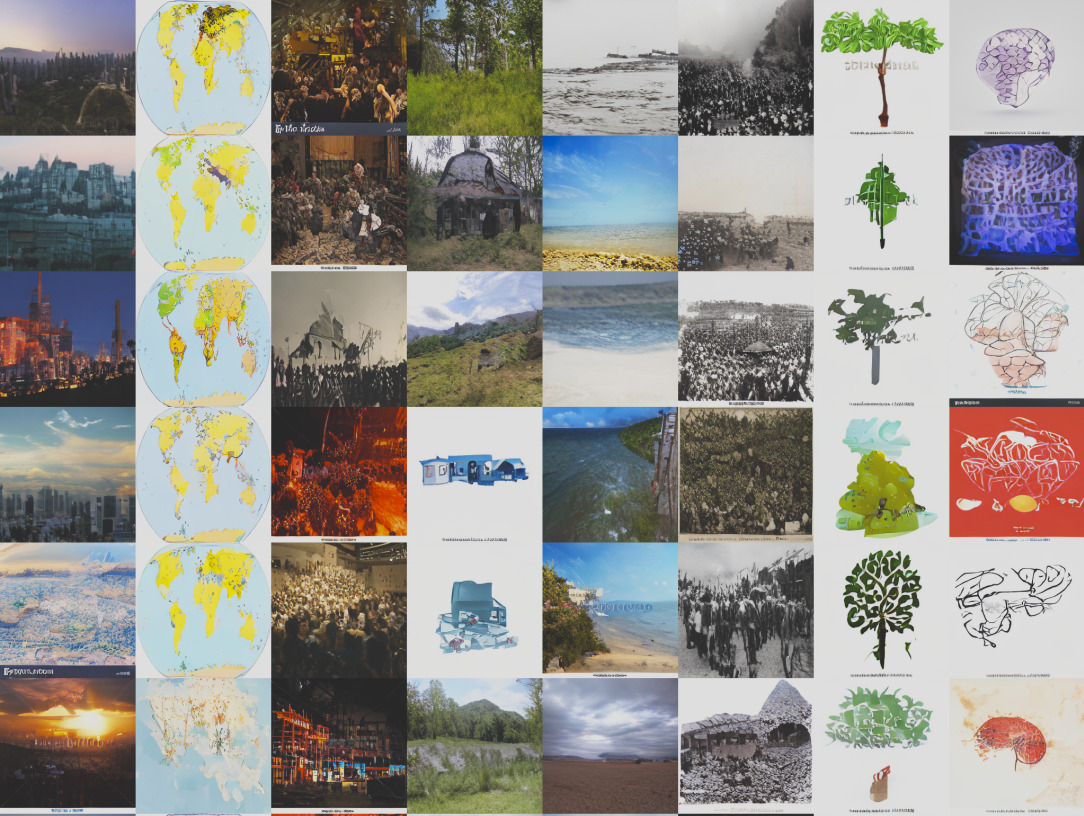}} \\
\end{tabular}
\begin{tabular}{m{\colwidth}m{\colwidth}m{\colwidth}m{\colwidth}m{\colwidth}m{\colwidth}m{\colwidth}m{\colwidth}}
\cmidrule[\cmidrulewidth]{1-8} 
\multicolumn{8}{c}{\small{TT~\citep{DBLP:journals/corr/abs-2012-09841}}} \\
\cmidrule[\cmidrulewidth]{1-8} 
\multicolumn{8}{c}{\includegraphics[width=.95\textwidth]{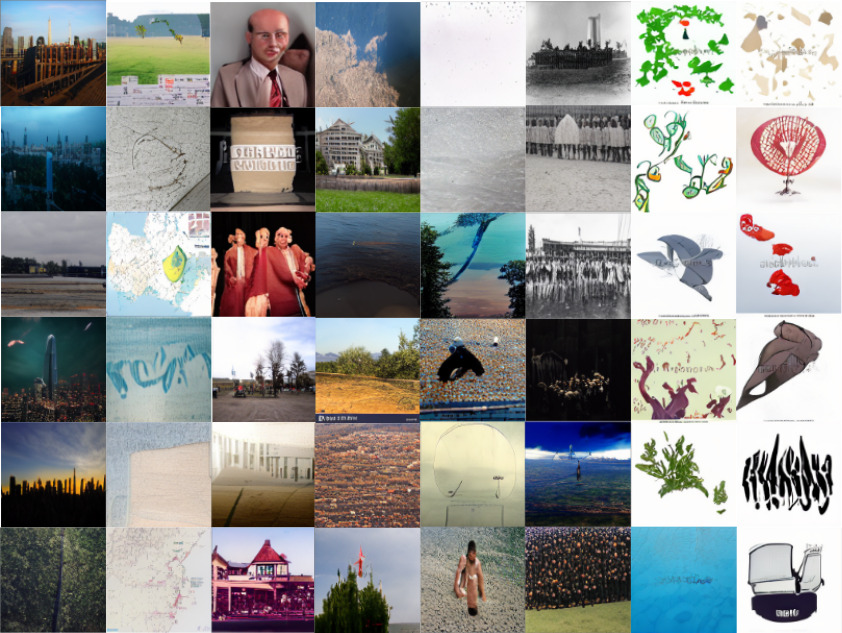}}
\end{tabular}
\end{footnotesize}
\caption{\label{fig:comptext2imgsupp} Random samples of text-conditional ImageBART and the text-conditional version of TT for the user defined text prompts above each row.}
\end{figure*}
}

\newcommand{\conceptualcaptionssupp}{
\begin{figure*}[thbp]
	\centering
    \renewcommand{\imwidth}{0.12\textwidth}
    \renewcommand{\impatha}[1]{img/conceptuals/heartmadeof/##1} 
    \renewcommand{\impathb}[1]{img/conceptuals/vasemadeof/##1} 
	\setlength{\tabcolsep}{0pt}
    \renewcommand{\arraystretch}{0}
	\begin{small}
 	\begin{tabular}{cccc|cccc}
 	\multicolumn{4}{c}{\small{A heart made of \emph{\_\_\_}.}} & \multicolumn{4}{c}{\small{A vase made of \emph{\_\_\_}.}} \\ 
  	\emph{\tiny{wood}} & \emph{\tiny{stone}} & \emph{\tiny{pizza}} & \emph{\tiny{metal}} & 
  	\emph{\tiny{wood}} & \emph{\tiny{avocado}} & \emph{\tiny{pizza}} & \emph{\tiny{pasta}} \\
  	\toprule
	\includegraphics[width=\imwidth]{\impatha{wood/clip_0003}} &	         
	\includegraphics[width=\imwidth]{\impatha{stone/subj_0001}} &	        
	\includegraphics[width=\imwidth]{\impatha{pizza/clip_0001}} &	         
	\includegraphics[width=\imwidth]{\impatha{metal/clip_0001}} &
	
	\includegraphics[width=\imwidth]{\impathb{wood/clip_0003}} &	         
	\includegraphics[width=\imwidth]{\impathb{avocado/clip_0001}} &	        
	\includegraphics[width=\imwidth]{\impathb{pizza/clip_0006}} &
	\includegraphics[width=\imwidth]{\impathb{pasta/clip_0001}} \\

	\includegraphics[width=\imwidth]{\impatha{wood/clip_0006}} &	         
	\includegraphics[width=\imwidth]{\impatha{stone/clip_0003}} &	        
	\includegraphics[width=\imwidth]{\impatha{pizza/clip_0002}} &	         
	\includegraphics[width=\imwidth]{\impatha{metal/clip_0002}} &

	\includegraphics[width=\imwidth]{\impathb{wood/clip_0002}} &	         
	\includegraphics[width=\imwidth]{\impathb{avocado/clip_0002}} &	        
	\includegraphics[width=\imwidth]{\impathb{pizza/clip_0002}} &
	\includegraphics[width=\imwidth]{\impathb{pasta/clip_0002}} \\
	
	\includegraphics[width=\imwidth]{\impatha{wood/clip_0001}} &	         
	\includegraphics[width=\imwidth]{\impatha{stone/clip_0002}} &	        
	\includegraphics[width=\imwidth]{\impatha{pizza/clip_0003}} &	         
	\includegraphics[width=\imwidth]{\impatha{metal/clip_0005}} &
	
	\includegraphics[width=\imwidth]{\impathb{wood/clip_0001}} &	         
	\includegraphics[width=\imwidth]{\impathb{avocado/clip_0006}} &	        
	\includegraphics[width=\imwidth]{\impathb{pizza/clip_0008}} &
	\includegraphics[width=\imwidth]{\impathb{pasta/clip_0003}} \\
	
	\includegraphics[width=\imwidth]{\impatha{wood/clip_0004}} &	         
	\includegraphics[width=\imwidth]{\impatha{stone/00000000_0008}} &	        
	\includegraphics[width=\imwidth]{\impatha{pizza/clip_0004}} &	         
	\includegraphics[width=\imwidth]{\impatha{metal/clip_0004}} &
	
	\includegraphics[width=\imwidth]{\impathb{wood/clip_0004}} &	         
	\includegraphics[width=\imwidth]{\impathb{avocado/clip_0005}} &	        
	\includegraphics[width=\imwidth]{\impathb{pizza/clip_0001}} & 
	\includegraphics[width=\imwidth]{\impathb{pasta/clip_0004}} \\
	
  	\bottomrule
	\end{tabular}
	\end{small}
	\caption{\label{fig:conceptualcaptionssupp} Additional samples from our text-conditional model.}
\end{figure*}	
}

\newcommand{\cinone}{
\begin{figure*}[thbp]
	\centering
    \renewcommand{\imwidth}{0.45\textwidth}
    \renewcommand{\impaths}[1]{img/cherries/imagenet/##1} 
    \setlength{\tabcolsep}{1pt}
    \renewcommand{\arraystretch}{1}
	\begin{small}
 	\begin{tabular}{cc}
  	\toprule
	      
	\includegraphics[width=\imwidth]{\impaths{finchgrid.jpg}} &
	\includegraphics[width=\imwidth]{\impaths{lorikeetgrid.jpg}} \\

	\includegraphics[width=\imwidth]{\impaths{anemonegrid.jpg}} &
	\includegraphics[width=\imwidth]{\impaths{tenchgrid.jpg}} \\
  	\bottomrule
	\end{tabular}
	\end{small}
	\caption{\label{fig:cinone} Additional class-conditional $256 \times 256$ random samples on ImageNet. Depicted classes are \emph{11: goldfinch} (top left),	\emph{90: lorikeet} (top right), \emph{108: sea anemone} (bottom left) and \emph{0: tench} (bottom right).}
\end{figure*}
}

\newcommand{\cintwo}{
\begin{figure*}[thbp]
	\centering
    \renewcommand{\imwidth}{0.45\textwidth}
    \renewcommand{\impaths}[1]{img/cherries/imagenet/##1} 
    \setlength{\tabcolsep}{1pt}
    \renewcommand{\arraystretch}{1}
	\begin{small}
 	\begin{tabular}{cc}
  	\toprule
	      
	\includegraphics[width=\imwidth]{\impaths{terriergrid.jpg}} &
	\includegraphics[width=\imwidth]{\impaths{geysirgrid.jpg}} \\

	\includegraphics[width=\imwidth]{\impaths{burgergrid.jpg}} &
	\includegraphics[width=\imwidth]{\impaths{shipgrid.jpg}} \\
  	\bottomrule
	\end{tabular}
	\end{small}
	\caption{\label{fig:cintwo} Additional class-conditional $256 \times 256$ random samples on ImageNet. Depicted classes are \emph{200: tibetian terrier} (top left),	\emph{974: geyser} (top right), \emph{933: cheeseburger} (bottom left) and \emph{510: container ship} (bottom right).}
\end{figure*}
}

\newcommand{\sflickronepage}{
\begin{figure}
	\centering
	\renewcommand{\impaths}[1]{img/conditional/sflckr/##1}
	\includegraphics[width=\textwidth]{\impaths{one_pager_semsyn_mask}}
	\includegraphics[width=\textwidth]{\impaths{one_pager_semsyn1}}        
	\includegraphics[width=\textwidth]{\impaths{one_pager_semsyn2}} 
	\includegraphics[width=\textwidth]{\impaths{one_pager_semsyn3}} 
	\caption{\label{fig:sflckronepage} Semantically guided samples from ImageBART conditionally trained semantic maps from the on the S-FLCKR dataset, similar to the one shown in the top row. Image size is $1024 \times 410$. }
\end{figure}
}

\newcommand{\nnffhq}{
\begin{figure}
	\centering
	\renewcommand{\impaths}[1]{img/nns/ffhq/##1}
	\includegraphics[width=\textwidth]{\impaths{example_41-10nns}}
	\includegraphics[width=\textwidth]{\impaths{example_43-10nns}}        
	\includegraphics[width=\textwidth]{\impaths{example_25-10nns}} 
	\includegraphics[width=\textwidth]{\impaths{example_35-10nns}} 
	\includegraphics[width=\textwidth]{\impaths{example_37-10nns}} 
	\includegraphics[width=\textwidth]{\impaths{example_21-10nns}} 
	\includegraphics[width=\textwidth]{\impaths{example_45-10nns}} 
	\includegraphics[width=\textwidth]{\impaths{example_49-10nns}} 
	\includegraphics[width=\textwidth]{\impaths{example_50-10nns}} 
	\caption{\label{fig:nnsffhq} Nearest neighbors to samples from ImageBART from the FFHQ train set measured by averaging over different feature layers of a VGG-16 trained on ImageNet. The first example in each row shows a generated sample from our model. The remaining ones depict the corresponding nearest neighbors in ascending order.}
\end{figure}
}

\newcommand{\nnchurches}{
\begin{figure}
	\centering
	\renewcommand{\impaths}[1]{img/nns/churches/##1}
	\includegraphics[width=\textwidth]{\impaths{example_12-10nns}}
	\includegraphics[width=\textwidth]{\impaths{example_13-10nns}}        
	\includegraphics[width=\textwidth]{\impaths{example_14-10nns}} 
	\includegraphics[width=\textwidth]{\impaths{example_16-10nns}} 
	\includegraphics[width=\textwidth]{\impaths{example_2-10nns}} 
	\includegraphics[width=\textwidth]{\impaths{example_6-10nns}} 
	\includegraphics[width=\textwidth]{\impaths{example_7-10nns}} 
	\includegraphics[width=\textwidth]{\impaths{example_20-10nns}} 
	\includegraphics[width=\textwidth]{\impaths{example_10-10nns}} 
	\caption{\label{fig:nnschurch} Nearest neighbors to samples from ImageBART from the LSUN-churches train set measured by averaging over different feature layers of a VGG-16 trained on ImageNet. The first example in each row shows a generated sample from our model. The remaining ones depict the corresponding nearest neighbors in ascending order.}
\end{figure}
}

\newcommand{\conditionalinpaintingsupp}{
\begin{figure*}[thbp]
\centering
    \renewcommand{\imwidth}{0.16\textwidth}
    \renewcommand{\impaths}[1]{img/editing/imagenet/##1}  
	\setlength{\tabcolsep}{0pt}
    \renewcommand{\arraystretch}{0}
	\begin{small}
 	\begin{tabular}{c@{\hskip 5pt}c@{\hskip 5pt}cccc}
 	Original & Masked & \multicolumn{4}{c}{Guidance} \\ 
  	\toprule
  	\tiny{\emph{Red fox (c277)}} & & \tiny{\emph{Brown Bear (c294)}} & \tiny{\emph{White Wolf (c270)}} & \tiny{\emph{Leopard (c289)}} & \tiny{\emph{Gazelle (c353)}} \\
  	\midrule
	\includegraphics[width=\imwidth]{\impaths{fox_standing/mask}} &	         
	\includegraphics[width=\imwidth]{\impaths{fox_standing/original}} &	 
	       
	\includegraphics[width=\imwidth]{\impaths{fox_standing/ursus}} &	         
	\includegraphics[width=\imwidth]{\impaths{fox_standing/polar_wolf}} &   
	\includegraphics[width=\imwidth]{\impaths{fox_standing/leopard}} &
	\includegraphics[width=\imwidth]{\impaths{fox_standing/gazelle}} \\
	\midrule
	\tiny{\emph{Lorikeet(c90)}} & & \tiny{\emph{Snow Leopard (c288)}} & \tiny{\emph{Doberman (c236)}} & \tiny{\emph{Leopard (c289)}} & \tiny{\emph{Arctic Fox (c279)}} \\
  	\midrule
	\includegraphics[width=\imwidth]{\impaths{parrot/original}} &	         
	\includegraphics[width=\imwidth]{\impaths{parrot/mask}} &	 
	       
	\includegraphics[width=\imwidth]{\impaths{parrot/snow_leopard}} &	         
	\includegraphics[width=\imwidth]{\impaths{parrot/dobberman}} &   
	\includegraphics[width=\imwidth]{\impaths{parrot/leopard}} &
	\includegraphics[width=\imwidth]{\impaths{parrot/arctic_fox}} \\
	\midrule
	\tiny{\emph{Pizza (c963)}} & & \tiny{\emph{Carbonara (c959)}} & \tiny{\emph{Meat loaf (c962)}} & \tiny{\emph{Broccoli (c937)}} & \tiny{\emph{Bell pepper (c945)}} \\
  	\midrule
	\includegraphics[width=\imwidth]{\impaths{pizza/original}} &	         
	\includegraphics[width=\imwidth]{\impaths{pizza/mask}} &	        

	\includegraphics[width=\imwidth]{\impaths{pizza/carbonara_962}} &	         
	\includegraphics[width=\imwidth]{\impaths{pizza/meat_loaf}} &	
	\includegraphics[width=\imwidth]{\impaths{pizza/broccoli}} &	         
	\includegraphics[width=\imwidth]{\impaths{pizza/bell_pepper}} \\
	
	\midrule
	\tiny{\emph{Granny Smith (c948)}} & & \tiny{\emph{Pizza (c963)}} & \tiny{\emph{Orange (c950)}} & \tiny{\emph{Volleyball(c890)}} & \tiny{\emph{Bell pepper (c945)}} \\
  	\midrule
	\includegraphics[width=\imwidth]{\impaths{apple/original}} &	         
	\includegraphics[width=\imwidth]{\impaths{apple/mask}} &	        

	\includegraphics[width=\imwidth]{\impaths{apple/pizza}} &	         
	\includegraphics[width=\imwidth]{\impaths{apple/orange}} &	
	\includegraphics[width=\imwidth]{\impaths{apple/volleyball}} &	         
	\includegraphics[width=\imwidth]{\impaths{apple/bell_pepper}} \\

  	\bottomrule
	\end{tabular}
	\end{small}
  
	\caption{\label{fig:supplocalcondediting}Conditionally guided inpainting results obtained from conditional ImageBART trained on the  ImageNet dataset.}
\end{figure*}	
}

\newcommand{\conditionalinpaintingsupptxt}{
\begin{figure*}[thbp]
\centering
    \renewcommand{\imwidth}{0.2\textwidth}
    \renewcommand{\impaths}[1]{img/editing/cc/##1}  
	\setlength{\tabcolsep}{0pt}
    \renewcommand{\arraystretch}{0}
	\begin{small}
 	\begin{tabular}{c@{\hskip 2pt}c@{\hskip 2pt}ccc}
 	Original & Masked & \multicolumn{3}{c}{Guidance} \\ 
  	\toprule
  	\tiny{\emph{'Man standing on a mountain.'}} & & \tiny{\emph{'Solar Eclipse.'}} & \tiny{\emph{'Sunrise.'}} & \tiny{\emph{'Moonlight'}}  \\
  	\midrule	
	\includegraphics[width=\imwidth]{\impaths{mountain/original}} &	         
	\includegraphics[width=\imwidth]{\impaths{mountain/mask}} &	        

	\includegraphics[width=\imwidth]{\impaths{mountain/solar_eclipse}} &	         
	\includegraphics[width=\imwidth]{\impaths{mountain/sunrise}} &	
	\includegraphics[width=\imwidth]{\impaths{mountain/moonlight}} \\
	
	\midrule	
	\tiny{\emph{'The piece of paper.'}} & & \tiny{\emph{'A pencil sketch.'}} & \tiny{\emph{'A forest behind the window.'}} & \tiny{\emph{'Oil painting of a cathedral.'}}  \\
  	\midrule	
		\includegraphics[width=\imwidth]{\impaths{image/original}} &	         
	\includegraphics[width=\imwidth]{\impaths{image/mask}} &	        

	\includegraphics[width=\imwidth]{\impaths{image/pencil}} &	         
	\includegraphics[width=\imwidth]{\impaths{image/forest}} &	
	\includegraphics[width=\imwidth]{\impaths{image/oil_painting_cathedral}} \\
  	\bottomrule
	\end{tabular}
	\end{small}

	\caption{\label{fig:supplocalcondeditingtext} Additional results on conditional inpainting obtained from conditional ImageBART trained on the  Conceptual Captions dataset.}
\end{figure*}	
}

\newcommand{\suppfoxchain}{
\begin{figure*}[thbp]
\centering
	\includegraphics[width=\textwidth]{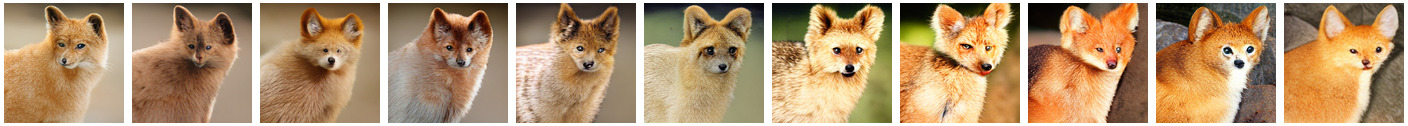}
	\caption{\label{fig:suppfoxchain}Running the forward-backward chain on a
  class conditional model allows for fine-grained exploration of samples for a
  given class, such as red fox (class 277).}
\end{figure*}	
}

\newcommand{\cincomp}{
\begin{figure*}[thbp]
	\centering
    \renewcommand{\imwidth}{0.08\textwidth}
    \renewcommand{\impaths}[1]{img/cincompare/##1} 
    \newcommand{\addimg}[1]{%
      \raisebox{-0.5\totalheight}{\includegraphics[width=\imwidth]{img/cincompare/##1}}}

    \setlength{\tabcolsep}{0pt}
    \renewcommand{\arraystretch}{1}
	\begin{small}
    \begin{tabular}{c@{\hskip 0.5em}cccccc@{\hskip 0.5em}cccccc}
  	\toprule
      rate & \multicolumn{6}{c}{ImageBART} & \multicolumn{6}{c}{TT
      \cite{DBLP:journals/corr/abs-2012-09841}} \\
      \midrule
      \multirow{6}{*}[-6.75em]{0.5} &
    \addimg{bart/0.5/sample_00} &
    \addimg{bart/0.5/sample_01} &
    \addimg{bart/0.5/sample_02} &
    \addimg{bart/0.5/sample_03} &
    \addimg{bart/0.5/sample_04} &
    \addimg{bart/0.5/sample_05} &
    \addimg{tt/0.5/sample_00} &
    \addimg{tt/0.5/sample_01} &
    \addimg{tt/0.5/sample_02} &
    \addimg{tt/0.5/sample_03} &
    \addimg{tt/0.5/sample_04} &
    \addimg{tt/0.5/sample_05} \\
    &
    \addimg{bart/0.5/sample_06} &
    \addimg{bart/0.5/sample_07} &
    \addimg{bart/0.5/sample_08} &
    \addimg{bart/0.5/sample_09} &
    \addimg{bart/0.5/sample_10} &
    \addimg{bart/0.5/sample_11} &
    \addimg{tt/0.5/sample_06} &
    \addimg{tt/0.5/sample_07} &
    \addimg{tt/0.5/sample_08} &
    \addimg{tt/0.5/sample_09} &
    \addimg{tt/0.5/sample_10} &
    \addimg{tt/0.5/sample_11} \\
    &
    \addimg{bart/0.5/sample_12} &
    \addimg{bart/0.5/sample_13} &
    \addimg{bart/0.5/sample_14} &
    \addimg{bart/0.5/sample_15} &
    \addimg{bart/0.5/sample_16} &
    \addimg{bart/0.5/sample_17} &
    \addimg{tt/0.5/sample_12} &
    \addimg{tt/0.5/sample_13} &
    \addimg{tt/0.5/sample_14} &
    \addimg{tt/0.5/sample_15} &
    \addimg{tt/0.5/sample_16} &
    \addimg{tt/0.5/sample_17} \\
    &
    \addimg{bart/0.5/sample_18} &
    \addimg{bart/0.5/sample_19} &
    \addimg{bart/0.5/sample_20} &
    \addimg{bart/0.5/sample_21} &
    \addimg{bart/0.5/sample_22} &
    \addimg{bart/0.5/sample_23} &
    \addimg{tt/0.5/sample_18} &
    \addimg{tt/0.5/sample_19} &
    \addimg{tt/0.5/sample_20} &
    \addimg{tt/0.5/sample_21} &
    \addimg{tt/0.5/sample_22} &
    \addimg{tt/0.5/sample_23} \\
    &
    \addimg{bart/0.5/sample_24} &
    \addimg{bart/0.5/sample_25} &
    \addimg{bart/0.5/sample_26} &
    \addimg{bart/0.5/sample_27} &
    \addimg{bart/0.5/sample_28} &
    \addimg{bart/0.5/sample_29} &
    \addimg{tt/0.5/sample_24} &
    \addimg{tt/0.5/sample_25} &
    \addimg{tt/0.5/sample_26} &
    \addimg{tt/0.5/sample_27} &
    \addimg{tt/0.5/sample_28} &
    \addimg{tt/0.5/sample_29} \\
    &
    \addimg{bart/0.5/sample_30} &
    \addimg{bart/0.5/sample_31} &
    \addimg{bart/0.5/sample_32} &
    \addimg{bart/0.5/sample_33} &
    \addimg{bart/0.5/sample_34} &
    \addimg{bart/0.5/sample_35} &
    \addimg{tt/0.5/sample_30} &
    \addimg{tt/0.5/sample_31} &
    \addimg{tt/0.5/sample_32} &
    \addimg{tt/0.5/sample_33} &
    \addimg{tt/0.5/sample_34} &
    \addimg{tt/0.5/sample_35} \\[1.75em]
    &
    \multicolumn{6}{c}{FID: 13.12} &
    \multicolumn{6}{c}{FID: 10.26} \\
    \midrule
      \multirow{6}{*}[-6.75em]{0.25} &
    \addimg{bart/0.25/sample_00} &
    \addimg{bart/0.25/sample_01} &
    \addimg{bart/0.25/sample_02} &
    \addimg{bart/0.25/sample_03} &
    \addimg{bart/0.25/sample_04} &
    \addimg{bart/0.25/sample_05} &
    \addimg{tt/0.25/sample_00} &
    \addimg{tt/0.25/sample_01} &
    \addimg{tt/0.25/sample_02} &
    \addimg{tt/0.25/sample_03} &
    \addimg{tt/0.25/sample_04} &
    \addimg{tt/0.25/sample_05} \\
    &
    \addimg{bart/0.25/sample_06} &
    \addimg{bart/0.25/sample_07} &
    \addimg{bart/0.25/sample_08} &
    \addimg{bart/0.25/sample_09} &
    \addimg{bart/0.25/sample_10} &
    \addimg{bart/0.25/sample_11} &
    \addimg{tt/0.25/sample_06} &
    \addimg{tt/0.25/sample_07} &
    \addimg{tt/0.25/sample_08} &
    \addimg{tt/0.25/sample_09} &
    \addimg{tt/0.25/sample_10} &
    \addimg{tt/0.25/sample_11} \\
    &
    \addimg{bart/0.25/sample_12} &
    \addimg{bart/0.25/sample_13} &
    \addimg{bart/0.25/sample_14} &
    \addimg{bart/0.25/sample_15} &
    \addimg{bart/0.25/sample_16} &
    \addimg{bart/0.25/sample_17} &
    \addimg{tt/0.25/sample_12} &
    \addimg{tt/0.25/sample_13} &
    \addimg{tt/0.25/sample_14} &
    \addimg{tt/0.25/sample_15} &
    \addimg{tt/0.25/sample_16} &
    \addimg{tt/0.25/sample_17} \\
    &
    \addimg{bart/0.25/sample_18} &
    \addimg{bart/0.25/sample_19} &
    \addimg{bart/0.25/sample_20} &
    \addimg{bart/0.25/sample_21} &
    \addimg{bart/0.25/sample_22} &
    \addimg{bart/0.25/sample_23} &
    \addimg{tt/0.25/sample_18} &
    \addimg{tt/0.25/sample_19} &
    \addimg{tt/0.25/sample_20} &
    \addimg{tt/0.25/sample_21} &
    \addimg{tt/0.25/sample_22} &
    \addimg{tt/0.25/sample_23} \\
    &
    \addimg{bart/0.25/sample_24} &
    \addimg{bart/0.25/sample_25} &
    \addimg{bart/0.25/sample_26} &
    \addimg{bart/0.25/sample_27} &
    \addimg{bart/0.25/sample_28} &
    \addimg{bart/0.25/sample_29} &
    \addimg{tt/0.25/sample_24} &
    \addimg{tt/0.25/sample_25} &
    \addimg{tt/0.25/sample_26} &
    \addimg{tt/0.25/sample_27} &
    \addimg{tt/0.25/sample_28} &
    \addimg{tt/0.25/sample_29} \\
    &
    \addimg{bart/0.25/sample_30} &
    \addimg{bart/0.25/sample_31} &
    \addimg{bart/0.25/sample_32} &
    \addimg{bart/0.25/sample_33} &
    \addimg{bart/0.25/sample_34} &
    \addimg{bart/0.25/sample_35} &
    \addimg{tt/0.25/sample_30} &
    \addimg{tt/0.25/sample_31} &
    \addimg{tt/0.25/sample_32} &
    \addimg{tt/0.25/sample_33} &
    \addimg{tt/0.25/sample_34} &
    \addimg{tt/0.25/sample_35} \\[1.75em]
    &
    \multicolumn{6}{c}{FID: 9.77} &
    \multicolumn{6}{c}{FID: 7.35} \\
    \midrule
      \multirow{6}{*}[-6.75em]{0.05} &
    \addimg{bart/0.05/sample_00} &
    \addimg{bart/0.05/sample_01} &
    \addimg{bart/0.05/sample_02} &
    \addimg{bart/0.05/sample_03} &
    \addimg{bart/0.05/sample_04} &
    \addimg{bart/0.05/sample_05} &
    \addimg{tt/0.05/sample_00} &
    \addimg{tt/0.05/sample_01} &
    \addimg{tt/0.05/sample_02} &
    \addimg{tt/0.05/sample_03} &
    \addimg{tt/0.05/sample_04} &
    \addimg{tt/0.05/sample_05} \\
    &
    \addimg{bart/0.05/sample_06} &
    \addimg{bart/0.05/sample_07} &
    \addimg{bart/0.05/sample_08} &
    \addimg{bart/0.05/sample_09} &
    \addimg{bart/0.05/sample_10} &
    \addimg{bart/0.05/sample_11} &
    \addimg{tt/0.05/sample_06} &
    \addimg{tt/0.05/sample_07} &
    \addimg{tt/0.05/sample_08} &
    \addimg{tt/0.05/sample_09} &
    \addimg{tt/0.05/sample_10} &
    \addimg{tt/0.05/sample_11} \\
    &
    \addimg{bart/0.05/sample_12} &
    \addimg{bart/0.05/sample_13} &
    \addimg{bart/0.05/sample_14} &
    \addimg{bart/0.05/sample_15} &
    \addimg{bart/0.05/sample_16} &
    \addimg{bart/0.05/sample_17} &
    \addimg{tt/0.05/sample_12} &
    \addimg{tt/0.05/sample_13} &
    \addimg{tt/0.05/sample_14} &
    \addimg{tt/0.05/sample_15} &
    \addimg{tt/0.05/sample_16} &
    \addimg{tt/0.05/sample_17} \\
    &
    \addimg{bart/0.05/sample_18} &
    \addimg{bart/0.05/sample_19} &
    \addimg{bart/0.05/sample_20} &
    \addimg{bart/0.05/sample_21} &
    \addimg{bart/0.05/sample_22} &
    \addimg{bart/0.05/sample_23} &
    \addimg{tt/0.05/sample_18} &
    \addimg{tt/0.05/sample_19} &
    \addimg{tt/0.05/sample_20} &
    \addimg{tt/0.05/sample_21} &
    \addimg{tt/0.05/sample_22} &
    \addimg{tt/0.05/sample_23} \\
    &
    \addimg{bart/0.05/sample_24} &
    \addimg{bart/0.05/sample_25} &
    \addimg{bart/0.05/sample_26} &
    \addimg{bart/0.05/sample_27} &
    \addimg{bart/0.05/sample_28} &
    \addimg{bart/0.05/sample_29} &
    \addimg{tt/0.05/sample_24} &
    \addimg{tt/0.05/sample_25} &
    \addimg{tt/0.05/sample_26} &
    \addimg{tt/0.05/sample_27} &
    \addimg{tt/0.05/sample_28} &
    \addimg{tt/0.05/sample_29} \\
    &
    \addimg{bart/0.05/sample_30} &
    \addimg{bart/0.05/sample_31} &
    \addimg{bart/0.05/sample_32} &
    \addimg{bart/0.05/sample_33} &
    \addimg{bart/0.05/sample_34} &
    \addimg{bart/0.05/sample_35} &
    \addimg{tt/0.05/sample_30} &
    \addimg{tt/0.05/sample_31} &
    \addimg{tt/0.05/sample_32} &
    \addimg{tt/0.05/sample_33} &
    \addimg{tt/0.05/sample_34} &
    \addimg{tt/0.05/sample_35} \\[1.75em]
    &
    \multicolumn{6}{c}{FID: 7.44} &
    \multicolumn{6}{c}{FID: 5.88} \\
  	\bottomrule
	\end{tabular}
	\end{small}
	\caption{\label{fig:cincomp} Qualitative and quantitative comparison of cIN samples for
  different rejection rates as in
  Tab.~\ref{tab:fids}.
  }
\end{figure*}
}

%% file: tables_supplementary.tex
\newcommand{\compressionhyper}{
\begin{table}[tbhp]
\centering
\begin{small}
\begin{tabular}{l c c c c c c}
experiment & section & \makecell{add. \\ samples} & \makecell{effective \\ size} & \makecell{fine-tuned \\ \footnotesize{from \citep{DBLP:journals/corr/abs-2012-09841}}} & \makecell{trained \\ \footnotesize{from scratch}} & \makecell{compression \\ rate} \\
\toprule
\emph{class-cond. ImageNet} & \ref{subsec:expone} & Fig.~\ref{fig:imagenetsamplessupp},\ref{fig:cinone},\ref{fig:cintwo} & 973 & \xmark & \xmark  & $\nicefrac{1}{256}$ \\
\midrule
\emph{LSUN-Cats} & \ref{subsec:expone} & Fig.~\ref{fig:catsgridsupp} & 1014 & \cmark & \xmark & $\nicefrac{1}{256}$ \\
\emph{LSUN-Churches} & \ref{subsec:expone} & Fig.~\ref{fig:churchessamplessupp} & 1022 & \cmark & \xmark & $\nicefrac{1}{256}$ \\
\emph{LSUN-Bedrooms} & \ref{subsec:expone} & Fig.~\ref{fig:bedroomsgridsupp} & 1017 & \cmark & \xmark & $\nicefrac{1}{256}$ \\
\midrule
\emph{Conceptual Captions} & \ref{subsec:exptwo} & Fig.~\ref{fig:conceptualcaptionssupp} & 973 & \xmark & \xmark & $\nicefrac{1}{256}$ \\
\midrule
\emph{FFHQ} & \ref{subsec:expone} & Fig.~\ref{fig:ffhqsamplessupp} & 548 & \xmark & \cmark & $\nicefrac{1}{256}$ \\
\midrule
\emph{Semantic FLICKR} & \ref{subsec:morecond} & Fig.~\ref{fig:sflckronepage},\ref{fig:sflickrsamplesone} &  973 & \xmark & \xmark & $\nicefrac{1}{256}$ \\
\bottomrule
\end{tabular}
\end{small}
\caption{\label{tab:compressionhyper} Hyperparameters for all compression models used in our experiments.}
\end{table}
}

\newcommand{\diffusionhyper}{
\begin{table}%
\centering
\begin{small}
\begin{tabular}{l c c c}
experiment & length of chain & $\beta_t$ schedule ($t \geq 2$) & \makecell{effect. seq. length \\ \scriptsize{(full $\seqlength=256$, $t \geq 2$, w/o cond.)}} \\
\toprule
\emph{class-cond. ImageNet} & $T=6$ & $[0.090, 0.104, 0.139, 0.266, 1.0]$ & $[232, 208, 179, 131, 0]$ \\
\midrule
\emph{LSUN-Cats} & $T=4$ & $[0.152, 0.231, 1.0]$ & $[217, 166, 0]$ \\
\emph{LSUN-Churches} & $T=4$ & $[0.152, 0.231, 1.0]$ & $[217, 166, 0]$ \\
\emph{LSUN-Bedrooms} &  $T=4$ & $[0.152, 0.231, 1.0]$ & $[217, 166, 0]$ \\
\midrule
\emph{Conceptual Captions} &  $T=5$ & $[0.113, 0.141, 0.246, 1.0]$ & $[227, 195, 147, 0]$ \\
\midrule
\emph{FFHQ} & $T=3$ & $[0.364, 1.0]$ & $[162, 0]$ \\
\midrule
\emph{Semantic FLICKR} & $T=5$ & $[0.250, 0.333, 0.500, 1.0]$ & $[192, 128, 64, 0]$ \\
\bottomrule
\end{tabular}
\end{small}
\caption{\label{tab:diffusionhyper} Hyperparameters for all multinomial diffusion process we used in our experiments.}
\end{table}
}

\newcommand{\transformerhyper}{
\begin{table}[tbhp]
\centering
\begin{footnotesize}
\begin{tabular}{l c c c}
experiment &  num. parameters [M] & \makecell{num. layers \\ \scriptsize{(encoder/decoder)}} & \makecell{embed. dim. \\ \scriptsize{(encoder \& decoder)}} \\
\toprule
\emph{class-cond. ImageNet} & $[693, 693, 693, 693, 718]$ & $[4\times (32/6), (0/36)]$ & $[4\times 1152, 1216]$ \\
\midrule
\emph{LSUN-Cats} & $[693, 693, 718]$ & $[2 \times (32/6), (0/36)]$ & $[2 \times 1152, 1216]$ \\
\emph{LSUN-Churches} & $[693, 693, 718]$ & $[2 \times (32/6), (0/36)]$ & $[2 \times 1152, 1216]$ \\
\emph{LSUN-Bedrooms} & $[693, 693, 718]$ & $[2 \times (32/6), (0/36)]$ & $[2 \times 1152, 1216]$ \\
\midrule
\emph{Conceptual Captions} & $[685, 685, 685, 778]$ & $[3 \times (32/6), (0/36)]$ & $[3\times 1152, 1216]$ \\
\midrule
\emph{FFHQ} & $[687, 713]$ & $[(32/6), (0/36)]$ & $[1152, 1216]$ \\
\midrule
\emph{Semantic FLICKR} & $[397,397,397,429]$ & $[3 \times (19/5),(0,24)]$ & $[1152,1216]$ \\
\bottomrule
\end{tabular}
\end{footnotesize}
\caption{\label{tab:transformerhyper} Hyperparameters for each experiment and scale ($t \geq 2$) used to implement the Markov chain in Eq.~\eqref{eq:mdiff}.}
\end{table}
}

%% file: ms.bbl
\begin{thebibliography}{78}
\providecommand{\natexlab}[1]{#1}
\providecommand{\url}[1]{\texttt{#1}}
\expandafter\ifx\csname urlstyle\endcsname\relax
  \providecommand{\doi}[1]{doi: #1}\else
  \providecommand{\doi}{doi: \begingroup \urlstyle{rm}\Url}\fi

\bibitem[Abdal et~al.(2021)Abdal, Zhu, Mitra, and Wonka]{10.1145/3447648}
R.~Abdal, P.~Zhu, N.~J. Mitra, and P.~Wonka.
\newblock Styleflow: Attribute-conditioned exploration of stylegan-generated
  images using conditional continuous normalizing flows.
\newblock \emph{ACM Trans. Graph.}, 40\penalty0 (3), May 2021.
\newblock ISSN 0730-0301.
\newblock \doi{10.1145/3447648}.
\newblock URL \url{https://doi.org/10.1145/3447648}.

\bibitem[Austin et~al.(2021)Austin, Johnson, Ho, Tarlow, and
  Berg]{austin2021structured}
J.~Austin, D.~Johnson, J.~Ho, D.~Tarlow, and R.~v.~d. Berg.
\newblock Structured denoising diffusion models in discrete state-spaces.
\newblock \emph{arXiv preprint arXiv:2107.03006}, 2021.

\bibitem[Bengio et~al.(2015)Bengio, Vinyals, Jaitly, and
  Shazeer]{DBLP:journals/corr/BengioVJS15}
S.~Bengio, O.~Vinyals, N.~Jaitly, and N.~Shazeer.
\newblock Scheduled sampling for sequence prediction with recurrent neural
  networks.
\newblock \emph{CoRR}, abs/1506.03099, 2015.

\bibitem[Brock et~al.(2019)Brock, Donahue, and Simonyan]{big_gan_brock}
A.~Brock, J.~Donahue, and K.~Simonyan.
\newblock Large scale {GAN} training for high fidelity natural image synthesis.
\newblock In \emph{Int. Conf. Learn. Represent.}, 2019.

\bibitem[Brown et~al.(2020)Brown, Mann, Ryder, Subbiah, Kaplan, Dhariwal,
  Neelakantan, Shyam, Sastry, Askell, Agarwal, Herbert{-}Voss, Krueger,
  Henighan, Child, Ramesh, Ziegler, Wu, Winter, Hesse, Chen, Sigler, Litwin,
  Gray, Chess, Clark, Berner, McCandlish, Radford, Sutskever, and
  Amodei]{DBLP:conf/nips/BrownMRSKDNSSAA20}
T.~B. Brown, B.~Mann, N.~Ryder, M.~Subbiah, J.~Kaplan, P.~Dhariwal,
  A.~Neelakantan, P.~Shyam, G.~Sastry, A.~Askell, S.~Agarwal,
  A.~Herbert{-}Voss, G.~Krueger, T.~Henighan, R.~Child, A.~Ramesh, D.~M.
  Ziegler, J.~Wu, C.~Winter, C.~Hesse, M.~Chen, E.~Sigler, M.~Litwin, S.~Gray,
  B.~Chess, J.~Clark, C.~Berner, S.~McCandlish, A.~Radford, I.~Sutskever, and
  D.~Amodei.
\newblock Language models are few-shot learners.
\newblock In \emph{NeurIPS}, 2020.

\bibitem[Chen et~al.(2020)Chen, Radford, Child, Wu, Jun, Luan, and
  Sutskever]{DBLP:conf/icml/ChenRC0JLS20}
M.~Chen, A.~Radford, R.~Child, J.~Wu, H.~Jun, D.~Luan, and I.~Sutskever.
\newblock Generative pretraining from pixels.
\newblock In \emph{{ICML}}, volume 119 of \emph{Proceedings of Machine Learning
  Research}, pages 1691--1703. {PMLR}, 2020.

\bibitem[Chen et~al.(2016)Chen, Kingma, Salimans, Duan, Dhariwal, Schulman,
  Sutskever, and Abbeel]{DBLP:journals/corr/ChenKSDDSSA16}
X.~Chen, D.~P. Kingma, T.~Salimans, Y.~Duan, P.~Dhariwal, J.~Schulman,
  I.~Sutskever, and P.~Abbeel.
\newblock Variational lossy autoencoder.
\newblock \emph{CoRR}, abs/1611.02731, 2016.

\bibitem[Child(2020)]{DBLP:journals/corr/abs-2011-10650}
R.~Child.
\newblock Very deep vaes generalize autoregressive models and can outperform
  them on images.
\newblock \emph{CoRR}, abs/2011.10650, 2020.

\bibitem[Crawford and Paglen(2019)]{crawford2019excavating}
K.~Crawford and T.~Paglen.
\newblock Excavating ai: The politics of training sets for machine learning,
  2019.
\newblock URL \url{https://excavating.ai}.

\bibitem[Dai and Wipf(2019)]{DBLP:conf/iclr/DaiW19}
B.~Dai and D.~P. Wipf.
\newblock Diagnosing and enhancing {VAE} models.
\newblock In \emph{{ICLR} (Poster)}. OpenReview.net, 2019.

\bibitem[Deng et~al.(2009)Deng, Dong, Socher, Li, Li, and
  Li]{DBLP:conf/cvpr/DengDSLL009}
J.~Deng, W.~Dong, R.~Socher, L.~Li, K.~Li, and F.~Li.
\newblock Imagenet: {A} large-scale hierarchical image database.
\newblock In \emph{{CVPR}}, pages 248--255. {IEEE} Computer Society, 2009.

\bibitem[Devlin et~al.(2018)Devlin, Chang, Lee, and
  Toutanova]{DBLP:journals/corr/abs-1810-04805}
J.~Devlin, M.~Chang, K.~Lee, and K.~Toutanova.
\newblock {BERT:} pre-training of deep bidirectional transformers for language
  understanding.
\newblock \emph{CoRR}, abs/1810.04805, 2018.

\bibitem[Dhariwal and Nichol(2021)]{DBLP:journals/corr/abs-2105-05233}
P.~Dhariwal and A.~Nichol.
\newblock Diffusion models beat gans on image synthesis.
\newblock \emph{CoRR}, abs/2105.05233, 2021.
\newblock URL \url{https://arxiv.org/abs/2105.05233}.

\bibitem[Dhariwal et~al.(2020)Dhariwal, Jun, Payne, Kim, Radford, and
  Sutskever]{DBLP:journals/corr/abs-2005-00341}
P.~Dhariwal, H.~Jun, C.~Payne, J.~W. Kim, A.~Radford, and I.~Sutskever.
\newblock Jukebox: {A} generative model for music.
\newblock \emph{CoRR}, abs/2005.00341, 2020.

\bibitem[Dieleman(2020)]{dieleman2020typicality}
S.~Dieleman.
\newblock Musings on typicality, 2020.
\newblock URL \url{https://benanne.github.io/2020/09/01/typicality.html}.

\bibitem[Dosovitskiy and Brox(2016)]{DBLP:conf/nips/DosovitskiyB16}
A.~Dosovitskiy and T.~Brox.
\newblock Generating images with perceptual similarity metrics based on deep
  networks.
\newblock In \emph{{NIPS}}, pages 658--666, 2016.

\bibitem[Esser et~al.(2020{\natexlab{a}})Esser, Rombach, and
  Ommer]{DBLP:journals/corr/abs-2004-13166}
P.~Esser, R.~Rombach, and B.~Ommer.
\newblock A disentangling invertible interpretation network for explaining
  latent representations.
\newblock \emph{CoRR}, abs/2004.13166, 2020{\natexlab{a}}.

\bibitem[Esser et~al.(2020{\natexlab{b}})Esser, Rombach, and
  Ommer]{DBLP:journals/corr/abs-2012-09841}
P.~Esser, R.~Rombach, and B.~Ommer.
\newblock Taming transformers for high-resolution image synthesis.
\newblock \emph{CoRR}, abs/2012.09841, 2020{\natexlab{b}}.

\bibitem[Fauw et~al.(2019)Fauw, Dieleman, and
  Simonyan]{DBLP:journals/corr/abs-1903-04933}
J.~D. Fauw, S.~Dieleman, and K.~Simonyan.
\newblock Hierarchical autoregressive image models with auxiliary decoders.
\newblock \emph{CoRR}, abs/1903.04933, 2019.

\bibitem[Gatys et~al.(2016)Gatys, Ecker, and Bethge]{DBLP:conf/cvpr/GatysEB16}
L.~A. Gatys, A.~S. Ecker, and M.~Bethge.
\newblock Image style transfer using convolutional neural networks.
\newblock In \emph{{CVPR}}, pages 2414--2423. {IEEE} Computer Society, 2016.

\bibitem[Germain et~al.(2015)Germain, Gregor, Murray, and
  Larochelle]{DBLP:journals/corr/GermainGML15}
M.~Germain, K.~Gregor, I.~Murray, and H.~Larochelle.
\newblock {MADE:} masked autoencoder for distribution estimation.
\newblock \emph{CoRR}, abs/1502.03509, 2015.

\bibitem[Goodfellow et~al.(2014)Goodfellow, Pouget{-}Abadie, Mirza, Xu,
  Warde{-}Farley, Ozair, Courville, and Bengio]{goodfellow2014GAN}
I.~J. Goodfellow, J.~Pouget{-}Abadie, M.~Mirza, B.~Xu, D.~Warde{-}Farley,
  S.~Ozair, A.~C. Courville, and Y.~Bengio.
\newblock Generative adversarial networks.
\newblock \emph{CoRR}, 2014.

\bibitem[Goyal et~al.(2016)Goyal, Lamb, Zhang, Zhang, Courville, and
  Bengio]{DBLP:conf/nips/GoyalLZZCB16}
A.~Goyal, A.~Lamb, Y.~Zhang, S.~Zhang, A.~C. Courville, and Y.~Bengio.
\newblock Professor forcing: {A} new algorithm for training recurrent networks.
\newblock In \emph{{NIPS}}, pages 4601--4609, 2016.

\bibitem[Gulrajani et~al.(2016)Gulrajani, Kumar, Ahmed, Ta{\"{\i}}ga, Visin,
  V{\'{a}}zquez, and Courville]{DBLP:journals/corr/GulrajaniKATVVC16}
I.~Gulrajani, K.~Kumar, F.~Ahmed, A.~A. Ta{\"{\i}}ga, F.~Visin,
  D.~V{\'{a}}zquez, and A.~C. Courville.
\newblock Pixelvae: {A} latent variable model for natural images.
\newblock \emph{CoRR}, abs/1611.05013, 2016.

\bibitem[Heusel et~al.(2017)Heusel, Ramsauer, Unterthiner, Nessler, and
  Hochreiter]{FID}
M.~Heusel, H.~Ramsauer, T.~Unterthiner, B.~Nessler, and S.~Hochreiter.
\newblock Gans trained by a two time-scale update rule converge to a local nash
  equilibrium.
\newblock In \emph{Adv. Neural Inform. Process. Syst.}, pages 6626--6637, 2017.

\bibitem[Ho et~al.(2020)Ho, Jain, and Abbeel]{DBLP:conf/nips/HoJA20}
J.~Ho, A.~Jain, and P.~Abbeel.
\newblock Denoising diffusion probabilistic models.
\newblock In \emph{NeurIPS}, 2020.

\bibitem[Hoogeboom et~al.(2021)Hoogeboom, Nielsen, Jaini, Forr{\'{e}}, and
  Welling]{DBLP:journals/corr/abs-2102-05379}
E.~Hoogeboom, D.~Nielsen, P.~Jaini, P.~Forr{\'{e}}, and M.~Welling.
\newblock Argmax flows and multinomial diffusion: Towards non-autoregressive
  language models.
\newblock \emph{CoRR}, abs/2102.05379, 2021.

\bibitem[Jha et~al.(2018)Jha, Anand, Singh, and
  Veeravasarapu]{DBLP:conf/eccv/JhaASV18}
A.~H. Jha, S.~Anand, M.~Singh, and V.~S.~R. Veeravasarapu.
\newblock Disentangling factors of variation with cycle-consistent variational
  auto-encoders.
\newblock In \emph{{ECCV} {(3)}}, volume 11207 of \emph{Lecture Notes in
  Computer Science}, pages 829--845. Springer, 2018.

\bibitem[Johnson et~al.(2016)Johnson, Alahi, and
  Fei{-}Fei]{DBLP:conf/eccv/JohnsonAF16}
J.~Johnson, A.~Alahi, and L.~Fei{-}Fei.
\newblock Perceptual losses for real-time style transfer and super-resolution.
\newblock In \emph{{ECCV} {(2)}}, volume 9906 of \emph{Lecture Notes in
  Computer Science}, pages 694--711. Springer, 2016.

\bibitem[Karras et~al.(2019{\natexlab{a}})Karras, Laine, and Aila]{stylegan}
T.~Karras, S.~Laine, and T.~Aila.
\newblock A style-based generator architecture for generative adversarial
  networks.
\newblock In \emph{2019 IEEE/CVF Conference on Computer Vision and Pattern
  Recognition (CVPR)}, 2019{\natexlab{a}}.

\bibitem[Karras et~al.(2019{\natexlab{b}})Karras, Laine, Aittala, Hellsten,
  Lehtinen, and Aila]{DBLP:journals/corr/abs-1912-04958}
T.~Karras, S.~Laine, M.~Aittala, J.~Hellsten, J.~Lehtinen, and T.~Aila.
\newblock Analyzing and improving the image quality of stylegan.
\newblock \emph{CoRR}, abs/1912.04958, 2019{\natexlab{b}}.

\bibitem[Kasai et~al.(2021)Kasai, Pappas, Peng, Cross, and
  Smith]{kasai2021deep}
J.~Kasai, N.~Pappas, H.~Peng, J.~Cross, and N.~A. Smith.
\newblock Deep encoder, shallow decoder: Reevaluating non-autoregressive
  machine translation, 2021.

\bibitem[Khan et~al.(2021)Khan, Naseer, Hayat, Zamir, Khan, and
  Shah]{DBLP:journals/corr/abs-2101-01169}
S.~Khan, M.~Naseer, M.~Hayat, S.~W. Zamir, F.~S. Khan, and M.~Shah.
\newblock Transformers in vision: {A} survey.
\newblock \emph{CoRR}, abs/2101.01169, 2021.

\bibitem[Kilcher et~al.(2017)Kilcher, Lucchi, and
  Hofmann]{DBLP:journals/corr/abs-1710-11381}
Y.~Kilcher, A.~Lucchi, and T.~Hofmann.
\newblock Semantic interpolation in implicit models.
\newblock \emph{CoRR}, abs/1710.11381, 2017.

\bibitem[Kingma and Welling(2014)]{VAE}
D.~P. Kingma and M.~Welling.
\newblock {Auto-Encoding Variational Bayes}.
\newblock In \emph{2nd International Conference on Learning Representations,
  {ICLR}}, 2014.

\bibitem[Kingma et~al.(2014)Kingma, Rezende, Mohamed, and
  Welling]{DBLP:journals/corr/KingmaRMW14}
D.~P. Kingma, D.~J. Rezende, S.~Mohamed, and M.~Welling.
\newblock Semi-supervised learning with deep generative models.
\newblock \emph{CoRR}, abs/1406.5298, 2014.

\bibitem[Kolesnikov and Lampert(2017)]{DBLP:conf/icml/KolesnikovL17}
A.~Kolesnikov and C.~H. Lampert.
\newblock Pixelcnn models with auxiliary variables for natural image modeling.
\newblock In \emph{{ICML}}, volume~70 of \emph{Proceedings of Machine Learning
  Research}, pages 1905--1914. {PMLR}, 2017.

\bibitem[Kolmogoroff(1931)]{Kolmogoroff1931}
A.~Kolmogoroff.
\newblock Über die analytischen methoden in der wahrscheinlichkeitsrechnung.
\newblock \emph{Mathematische Annalen}, 104:\penalty0 415--458, 1931.
\newblock URL \url{http://eudml.org/doc/159476}.

\bibitem[Krizhevsky(2009)]{Krizhevsky2009LearningML}
A.~Krizhevsky.
\newblock Learning multiple layers of features from tiny images.
\newblock 2009.

\bibitem[Leblond et~al.(2017)Leblond, Alayrac, Osokin, and
  Lacoste{-}Julien]{DBLP:journals/corr/LeblondAOL17}
R.~Leblond, J.~Alayrac, A.~Osokin, and S.~Lacoste{-}Julien.
\newblock {SEARNN:} training rnns with global-local losses.
\newblock \emph{CoRR}, abs/1706.04499, 2017.

\bibitem[Lesniak et~al.(2019)Lesniak, Sieradzki, and
  Podolak]{DBLP:conf/iclr/LesniakSP19}
D.~Lesniak, I.~Sieradzki, and I.~T. Podolak.
\newblock Distribution-interpolation trade off in generative models.
\newblock In \emph{{ICLR} (Poster)}. OpenReview.net, 2019.

\bibitem[Lewis et~al.(2020)Lewis, Liu, Goyal, Ghazvininejad, Mohamed, Levy,
  Stoyanov, and Zettlemoyer]{DBLP:conf/acl/LewisLGGMLSZ20}
M.~Lewis, Y.~Liu, N.~Goyal, M.~Ghazvininejad, A.~Mohamed, O.~Levy, V.~Stoyanov,
  and L.~Zettlemoyer.
\newblock {BART:} denoising sequence-to-sequence pre-training for natural
  language generation, translation, and comprehension.
\newblock In \emph{{ACL}}, pages 7871--7880. Association for Computational
  Linguistics, 2020.

\bibitem[Li et~al.(2019)Li, Liu, Liu, Zhao, and Liu]{DBLP:conf/aaai/Li0LZL19}
N.~Li, S.~Liu, Y.~Liu, S.~Zhao, and M.~Liu.
\newblock Neural speech synthesis with transformer network.
\newblock In \emph{{AAAI}}, pages 6706--6713. {AAAI} Press, 2019.

\bibitem[Maal{\o}e et~al.(2019)Maal{\o}e, Fraccaro, Li{\'{e}}vin, and
  Winther]{DBLP:conf/nips/MaaloeFLW19}
L.~Maal{\o}e, M.~Fraccaro, V.~Li{\'{e}}vin, and O.~Winther.
\newblock {BIVA:} {A} very deep hierarchy of latent variables for generative
  modeling.
\newblock In \emph{NeurIPS}, pages 6548--6558, 2019.

\bibitem[Mathieu et~al.(2016)Mathieu, Zhao, Sprechmann, Ramesh, and
  LeCun]{DBLP:journals/corr/MathieuZSRL16}
M.~Mathieu, J.~J. Zhao, P.~Sprechmann, A.~Ramesh, and Y.~LeCun.
\newblock Disentangling factors of variation in deep representations using
  adversarial training.
\newblock \emph{CoRR}, abs/1611.03383, 2016.

\bibitem[Mentzer et~al.(2020)Mentzer, Toderici, Tschannen, and
  Agustsson]{DBLP:conf/nips/MentzerTTA20}
F.~Mentzer, G.~Toderici, M.~Tschannen, and E.~Agustsson.
\newblock High-fidelity generative image compression.
\newblock In \emph{NeurIPS}, 2020.

\bibitem[Nash et~al.(2021)Nash, Menick, Dieleman, and
  Battaglia]{DBLP:journals/corr/abs-2103-03841}
C.~Nash, J.~Menick, S.~Dieleman, and P.~W. Battaglia.
\newblock Generating images with sparse representations.
\newblock \emph{CoRR}, abs/2103.03841, 2021.

\bibitem[Ng et~al.(2020)Ng, Pang, Sharma, and Soricut]{ng2020understanding}
E.~G. Ng, B.~Pang, P.~Sharma, and R.~Soricut.
\newblock Understanding guided image captioning performance across domains.
\newblock \emph{arXiv preprint arXiv:2012.02339}, 2020.

\bibitem[Nichol and Dhariwal(2021)]{DBLP:journals/corr/abs-2102-09672}
A.~Nichol and P.~Dhariwal.
\newblock Improved denoising diffusion probabilistic models.
\newblock \emph{CoRR}, abs/2102.09672, 2021.

\bibitem[Park et~al.(2019)Park, Liu, Wang, and Zhu]{spade}
T.~Park, M.-Y. Liu, T.-C. Wang, and J.-Y. Zhu.
\newblock Semantic image synthesis with spatially-adaptive normalization.
\newblock In \emph{Proceedings of the IEEE Conference on Computer Vision and
  Pattern Recognition}, 2019.

\bibitem[Parmar et~al.(2018)Parmar, Vaswani, Uszkoreit, Kaiser, Shazeer, Ku,
  and Tran]{DBLP:conf/icml/ParmarVUKSKT18}
N.~Parmar, A.~Vaswani, J.~Uszkoreit, L.~Kaiser, N.~Shazeer, A.~Ku, and D.~Tran.
\newblock Image transformer.
\newblock In \emph{{ICML}}, volume~80 of \emph{Proceedings of Machine Learning
  Research}, pages 4052--4061. {PMLR}, 2018.

\bibitem[Radford et~al.(2021)Radford, Kim, Hallacy, Ramesh, Goh, Agarwal,
  Sastry, Askell, Mishkin, Clark, Krueger, and
  Sutskever]{DBLP:journals/corr/abs-2103-00020}
A.~Radford, J.~W. Kim, C.~Hallacy, A.~Ramesh, G.~Goh, S.~Agarwal, G.~Sastry,
  A.~Askell, P.~Mishkin, J.~Clark, G.~Krueger, and I.~Sutskever.
\newblock Learning transferable visual models from natural language
  supervision.
\newblock \emph{CoRR}, abs/2103.00020, 2021.

\bibitem[Ramesh et~al.(2021)Ramesh, Pavlov, Goh, Gray, Voss, Radford, Chen, and
  Sutskever]{DBLP:journals/corr/abs-2102-12092}
A.~Ramesh, M.~Pavlov, G.~Goh, S.~Gray, C.~Voss, A.~Radford, M.~Chen, and
  I.~Sutskever.
\newblock Zero-shot text-to-image generation.
\newblock \emph{CoRR}, abs/2102.12092, 2021.

\bibitem[Ranzato et~al.(2016)Ranzato, Chopra, Auli, and
  Zaremba]{DBLP:journals/corr/RanzatoCAZ15}
M.~Ranzato, S.~Chopra, M.~Auli, and W.~Zaremba.
\newblock Sequence level training with recurrent neural networks.
\newblock In \emph{{ICLR} (Poster)}, 2016.

\bibitem[Razavi et~al.(2019)Razavi, van~den Oord, and
  Vinyals]{DBLP:conf/nips/RazaviOV19}
A.~Razavi, A.~van~den Oord, and O.~Vinyals.
\newblock Generating diverse high-fidelity images with {VQ-VAE-2}.
\newblock In \emph{NeurIPS}, pages 14837--14847, 2019.

\bibitem[Rezende et~al.(2014)Rezende, Mohamed, and Wierstra]{VAE2}
D.~J. Rezende, S.~Mohamed, and D.~Wierstra.
\newblock {Stochastic backpropagation and approximate inference in deep
  generative models}.
\newblock In \emph{Proceedings of the 31st International Conference on
  International Conference on Machine Learning, ICML}, 2014.

\bibitem[Rombach et~al.(2020)Rombach, Esser, and
  Ommer]{DBLP:conf/nips/RombachEO20}
R.~Rombach, P.~Esser, and B.~Ommer.
\newblock Network-to-network translation with conditional invertible neural
  networks.
\newblock In \emph{NeurIPS}, 2020.

\bibitem[Salimans et~al.(2016)Salimans, Goodfellow, Zaremba, Cheung, Radford,
  and Chen]{Salimans2016ImprovedTF}
T.~Salimans, I.~Goodfellow, W.~Zaremba, V.~Cheung, A.~Radford, and X.~Chen.
\newblock Improved techniques for training gans.
\newblock In \emph{NIPS}, 2016.

\bibitem[Salimans et~al.(2017)Salimans, Karpathy, Chen, and
  Kingma]{DBLP:journals/corr/SalimansKCK17}
T.~Salimans, A.~Karpathy, X.~Chen, and D.~P. Kingma.
\newblock Pixelcnn++: Improving the pixelcnn with discretized logistic mixture
  likelihood and other modifications.
\newblock \emph{CoRR}, abs/1701.05517, 2017.

\bibitem[Schmidt(2019)]{DBLP:conf/emnlp/Schmidt19}
F.~Schmidt.
\newblock Generalization in generation: {A} closer look at exposure bias.
\newblock In \emph{NGT@EMNLP-IJCNLP}, pages 157--167. Association for
  Computational Linguistics, 2019.

\bibitem[Sharma et~al.(2018)Sharma, Ding, Goodman, and
  Soricut]{sharma2018conceptual}
P.~Sharma, N.~Ding, S.~Goodman, and R.~Soricut.
\newblock Conceptual captions: A cleaned, hypernymed, image alt-text dataset
  for automatic image captioning.
\newblock In \emph{Proceedings of ACL}, 2018.

\bibitem[Sohl{-}Dickstein et~al.(2015)Sohl{-}Dickstein, Weiss, Maheswaranathan,
  and Ganguli]{DBLP:journals/corr/Sohl-DicksteinW15}
J.~Sohl{-}Dickstein, E.~A. Weiss, N.~Maheswaranathan, and S.~Ganguli.
\newblock Deep unsupervised learning using nonequilibrium thermodynamics.
\newblock \emph{CoRR}, abs/1503.03585, 2015.

\bibitem[S{\o}nderby et~al.(2016)S{\o}nderby, Raiko, Maal{\o}e, S{\o}nderby,
  and Winther]{DBLP:conf/nips/SonderbyRMSW16}
C.~K. S{\o}nderby, T.~Raiko, L.~Maal{\o}e, S.~K. S{\o}nderby, and O.~Winther.
\newblock Ladder variational autoencoders.
\newblock In \emph{{NIPS}}, pages 3738--3746, 2016.

\bibitem[Song and Ermon(2019)]{DBLP:conf/nips/SongE19}
Y.~Song and S.~Ermon.
\newblock Generative modeling by estimating gradients of the data distribution.
\newblock In \emph{NeurIPS}, pages 11895--11907, 2019.

\bibitem[Song et~al.(2020)Song, Sohl{-}Dickstein, Kingma, Kumar, Ermon, and
  Poole]{DBLP:journals/corr/abs-2011-13456}
Y.~Song, J.~Sohl{-}Dickstein, D.~P. Kingma, A.~Kumar, S.~Ermon, and B.~Poole.
\newblock Score-based generative modeling through stochastic differential
  equations.
\newblock \emph{CoRR}, abs/2011.13456, 2020.

\bibitem[Szab{\'{o}} et~al.(2018)Szab{\'{o}}, Hu, Portenier, Zwicker, and
  Favaro]{DBLP:conf/iclr/SzaboHPZF18}
A.~Szab{\'{o}}, Q.~Hu, T.~Portenier, M.~Zwicker, and P.~Favaro.
\newblock Challenges in disentangling independent factors of variation.
\newblock In \emph{{ICLR} (Workshop)}. OpenReview.net, 2018.

\bibitem[Uria et~al.(2016)Uria, C{\^{o}}t{\'{e}}, Gregor, Murray, and
  Larochelle]{DBLP:journals/corr/UriaCGML16}
B.~Uria, M.~C{\^{o}}t{\'{e}}, K.~Gregor, I.~Murray, and H.~Larochelle.
\newblock Neural autoregressive distribution estimation.
\newblock \emph{CoRR}, abs/1605.02226, 2016.

\bibitem[Vahdat and Kautz(2020)]{DBLP:conf/nips/VahdatK20}
A.~Vahdat and J.~Kautz.
\newblock {NVAE:} {A} deep hierarchical variational autoencoder.
\newblock In \emph{NeurIPS}, 2020.

\bibitem[van~den Oord et~al.(2016{\natexlab{a}})van~den Oord, Kalchbrenner,
  Espeholt, kavukcuoglu, Vinyals, and Graves]{NIPS2016_b1301141}
A.~van~den Oord, N.~Kalchbrenner, L.~Espeholt, k.~kavukcuoglu, O.~Vinyals, and
  A.~Graves.
\newblock Conditional image generation with pixelcnn decoders.
\newblock In \emph{Advances in Neural Information Processing Systems},
  2016{\natexlab{a}}.

\bibitem[van~den Oord et~al.(2016{\natexlab{b}})van~den Oord, Kalchbrenner,
  Vinyals, Espeholt, Graves, and Kavukcuoglu]{DBLP:journals/corr/OordKVEGK16}
A.~van~den Oord, N.~Kalchbrenner, O.~Vinyals, L.~Espeholt, A.~Graves, and
  K.~Kavukcuoglu.
\newblock Conditional image generation with pixelcnn decoders.
\newblock \emph{CoRR}, abs/1606.05328, 2016{\natexlab{b}}.

\bibitem[van~den Oord et~al.(2017)van~den Oord, Vinyals, and
  Kavukcuoglu]{DBLP:conf/nips/OordVK17}
A.~van~den Oord, O.~Vinyals, and K.~Kavukcuoglu.
\newblock Neural discrete representation learning.
\newblock In \emph{{NIPS}}, pages 6306--6315, 2017.

\bibitem[Vaswani et~al.(2017)Vaswani, Shazeer, Parmar, Uszkoreit, Jones, Gomez,
  Kaiser, and Polosukhin]{DBLP:conf/nips/VaswaniSPUJGKP17}
A.~Vaswani, N.~Shazeer, N.~Parmar, J.~Uszkoreit, L.~Jones, A.~N. Gomez,
  L.~Kaiser, and I.~Polosukhin.
\newblock Attention is all you need.
\newblock In \emph{{NIPS}}, pages 5998--6008, 2017.

\bibitem[Wang et~al.(2020)Wang, Tang, Ma, Wu, Okhonko, and
  Pino]{wang2020fairseq}
C.~Wang, Y.~Tang, X.~Ma, A.~Wu, D.~Okhonko, and J.~Pino.
\newblock fairseq s2t: Fast speech-to-text modeling with fairseq, 2020.

\bibitem[White(2016)]{DBLP:journals/corr/White16a}
T.~White.
\newblock Sampling generative networks: Notes on a few effective techniques.
\newblock \emph{CoRR}, abs/1609.04468, 2016.

\bibitem[Yan et~al.(2021)Yan, Zhang, Abbeel, and
  Srinivas]{DBLP:journals/corr/abs-2104-10157}
W.~Yan, Y.~Zhang, P.~Abbeel, and A.~Srinivas.
\newblock Videogpt: Video generation using {VQ-VAE} and transformers.
\newblock \emph{CoRR}, abs/2104.10157, 2021.

\bibitem[Yu et~al.(2015)Yu, Zhang, Song, Seff, and
  Xiao]{DBLP:journals/corr/YuZSSX15}
F.~Yu, Y.~Zhang, S.~Song, A.~Seff, and J.~Xiao.
\newblock {LSUN:} construction of a large-scale image dataset using deep
  learning with humans in the loop.
\newblock \emph{CoRR}, abs/1506.03365, 2015.

\bibitem[Zhang et~al.(2018)Zhang, Isola, Efros, Shechtman, and Wang]{lpips}
R.~Zhang, P.~Isola, A.~A. Efros, E.~Shechtman, and O.~Wang.
\newblock The unreasonable effectiveness of deep features as a perceptual
  metric.
\newblock In \emph{Proceedings of the IEEE Conference on Computer Vision and
  Pattern Recognition (CVPR)}, June 2018.

\bibitem[Zheng et~al.(2021)Zheng, Cham, and
  Cai]{DBLP:journals/corr/abs-2104-00845}
C.~Zheng, T.~Cham, and J.~Cai.
\newblock Tfill: Image completion via a transformer-based architecture.
\newblock \emph{CoRR}, abs/2104.00845, 2021.

\end{thebibliography}
